\documentclass[twocolumn]{svjour3}          
\smartqed  
\usepackage{graphicx}
\usepackage{amsmath,amsfonts,amssymb}
\usepackage{subfigure}
\usepackage{natbib}
\usepackage{algorithm,algorithmic}
\usepackage{lineno}
\usepackage{color}
\usepackage{url}

\newcommand{\x}{\bf x}
\newcommand{\y}{\bf y}

\newcommand{\etal}{{\em et al.\/}\,}
\newcommand{\sgn}{\mathrm{sgn}}
\newcommand{\tr}{\mathrm{tr}}

\newcommand{\diag}{\mathrm{diag}}

\newcommand{\argmin}{\mathop{\rm argmin}}

\journalname{International Journal of Computer Vision}

\begin{document}

\title{Linearized Alternating Direction Method with Adaptive Penalty
and Warm Starts for Fast Solving Transform Invariant Low-Rank
Textures}

\author{Xiang Ren        \and
        Zhouchen Lin
}

\institute{Xiang Ren \at
              Department of Computer Science, University of Illinois at Urbana-Champaign, Urbana 61801, IL, USA \\
              \email{xren7@illinois.com}
           \and
           Z. Lin (corresponding author) \at
              Key Laboratory of Machine Perception (MOE), School of EECS, Peking University, Beijing 100871, P. R. China\\
              \email{zlin@pku.edu.cn}
}

\date{Received: date / Accepted: date}

\maketitle

\begin{abstract}
Transform Invariant Low-rank Textures (TILT) is a novel and
powerful tool that can effectively rectify a rich class of
low-rank textures in 3D scenes from 2D images despite significant
deformation and corruption. The existing algorithm for solving
TILT is based on the alternating direction method (ADM). It
suffers from high computational cost and is not theoretically
guaranteed to converge to a correct solution to the inner loop. In
this paper, we propose a novel algorithm to speed up solving TILT,
with guaranteed convergence for the inner loop. Our method is
based on the recently proposed linearized alternating direction
method with adaptive penalty (LADMAP). To further reduce
computation, warm starts are also introduced to initialize the
variables better and cut the cost on singular value decomposition.
Extensive experimental results on both synthetic and real data
demonstrate that this new algorithm works much more efficiently
and robustly than the existing algorithm. It could be at least
\emph{five times faster} than the previous method.
\end{abstract}

\section{Introduction}
\begin{figure}[t]
\begin{center}
\begin{tabular}{cc}
\includegraphics[width=0.45\columnwidth]{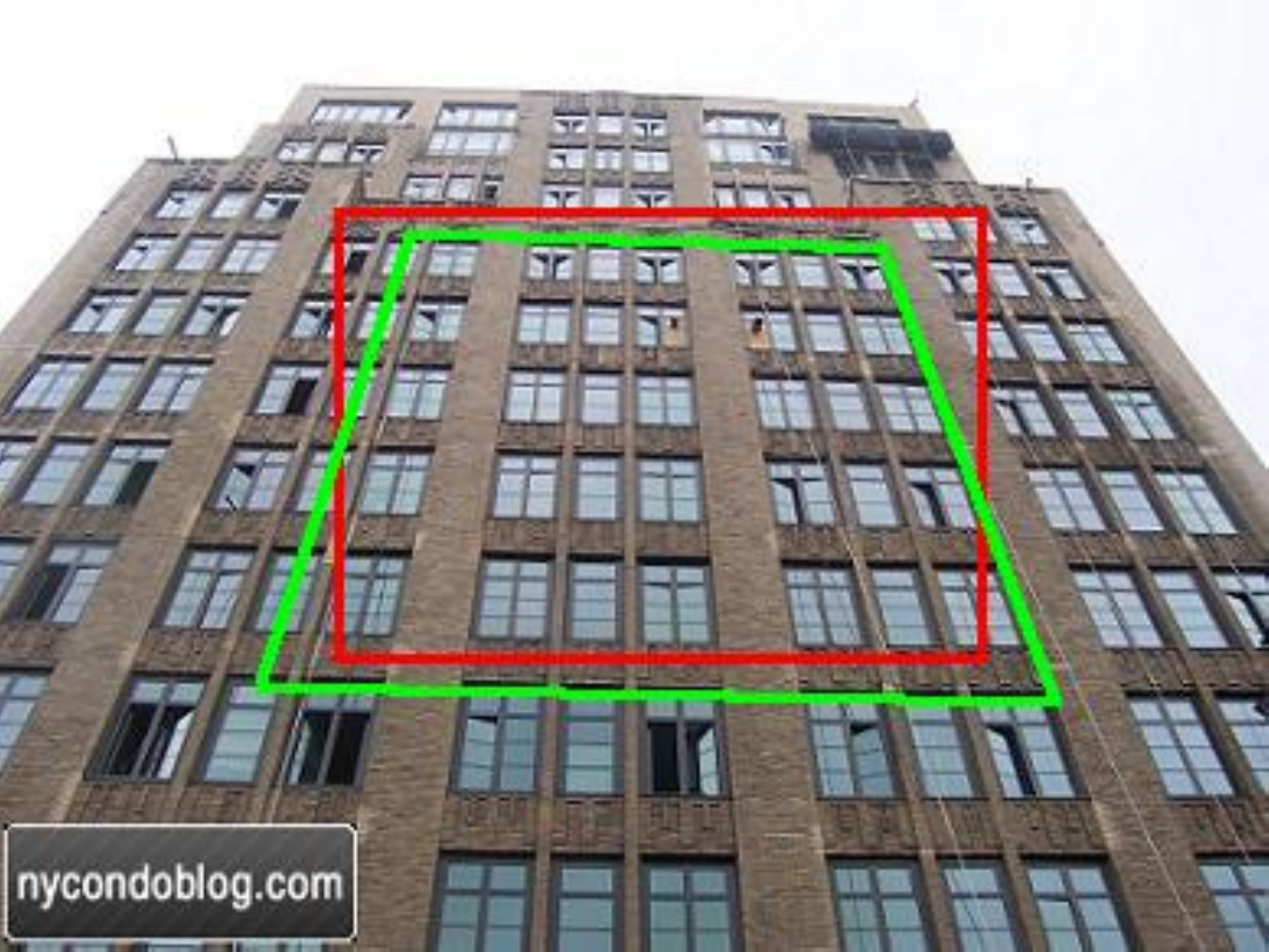}
&
\includegraphics[width=0.45\columnwidth]{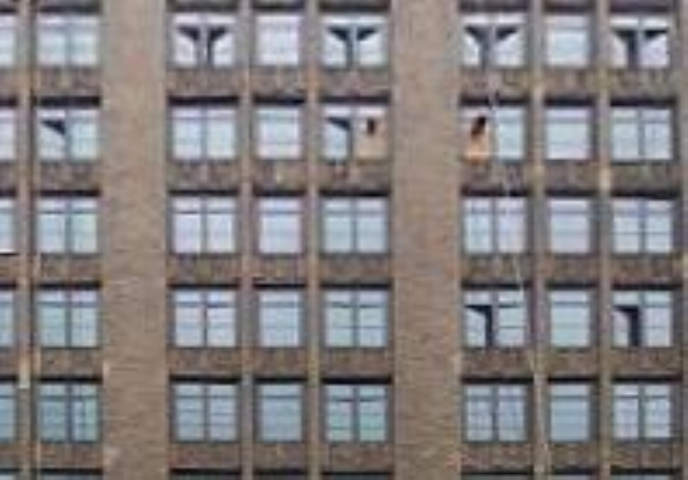}
\\
(a) & (b)\\
\includegraphics[width=0.45\columnwidth]{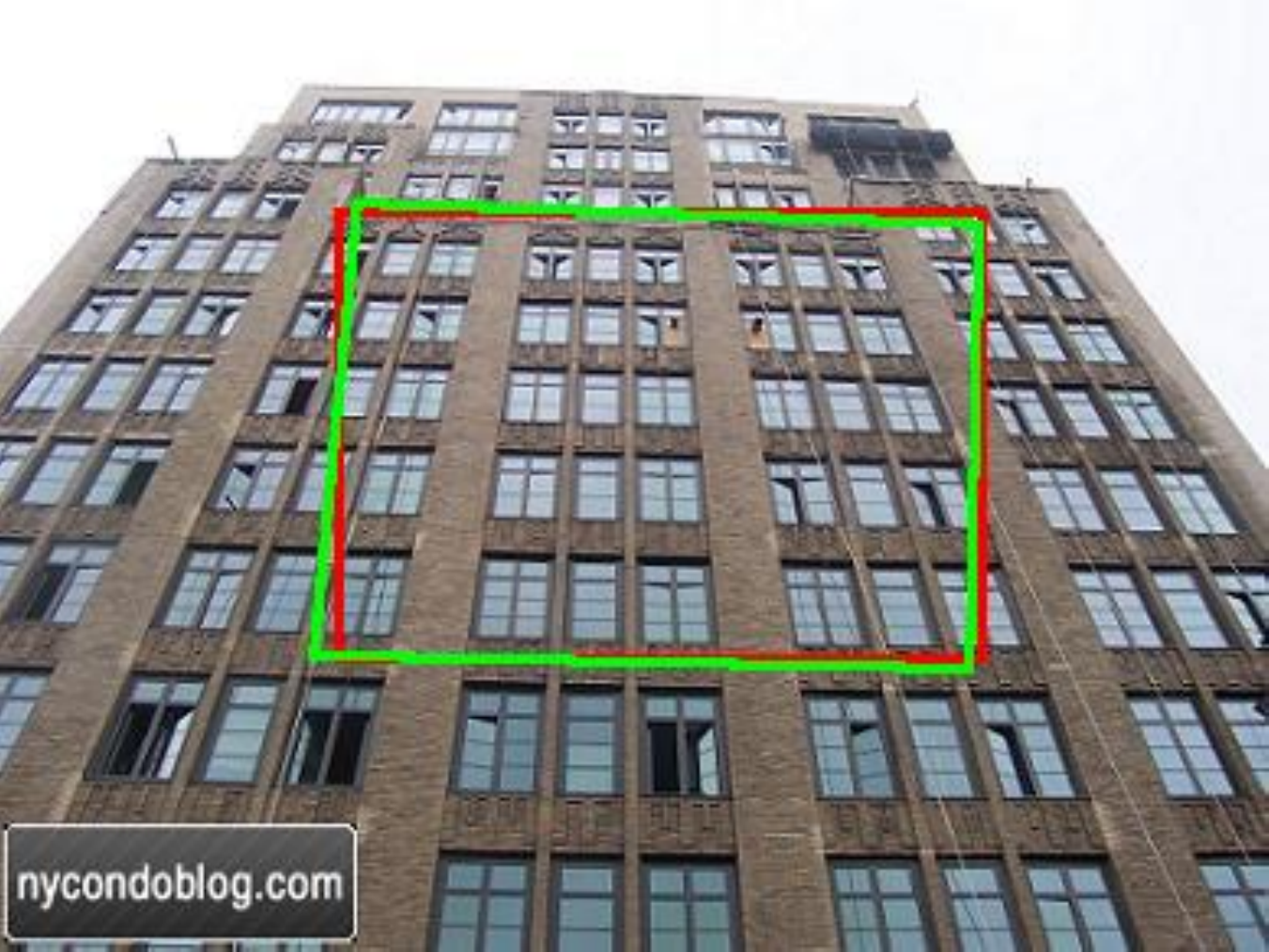}
&\includegraphics[width=0.45\columnwidth]{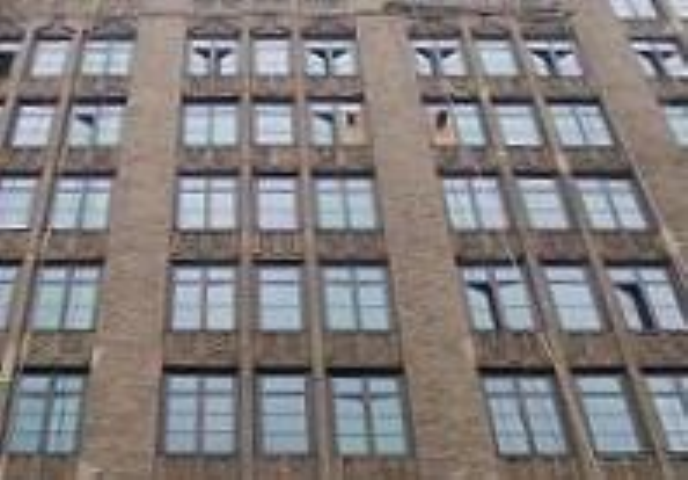}
\\
(c) & (d)
\end{tabular}
\caption{{\bf TILT for rectifying textures.} Left column: The
original image patches, specified by red rectangles, and the
transforms found by TILT, indicated by green quadrilaterals. Right
column: rectified textures using the transforms found. Top row:
results by our method for solving TILT. Bottom row: results by the
original method in \citep{ZhangTILT:11} for solving TILT. The
original method does not converge to a correct solution. {\bf
(Images in this paper are best viewed on screen!)}}
\label{fig:representative}
\end{center}
\end{figure}

Extracting invariants in images is a fundamental problem in
computer vision. It is key to many vision tasks, such as 3D
reconstruction, object recognition, and scene understanding. Most
of the existing methods start from low level local features, such
as SIFT points \citep{SIFT:2008}, corners, edges, and small
windows, which are inaccurate and unrobust. Recently, Zhang \etal
\citep{ZhangTILT:11} proposed a holistic method called the
Transform Invariant Low-rank Textures (TILT) that can recover the
deformation of a relatively large image patch so that the
underlying textures become regular (see
Figures~\ref{fig:representative} (a) and (b)). This method utilizes
the global structure in the image patch, e.g. the regularity, such
as symmetry, rectilinearity and repetition, that can be measured
by low rankness, rather than the low level local features, hence
can be very robust to significant deformation and corruption.

TILT has been applied to many vision tasks and been extended for
many computer vision applications. For example, Zhang
\etal~\citep{ZhangCVPR:11} used TILT to correct lens distortion
and do camera calibration, without detecting feature points and
straight lines. Zhang \etal~\citep{ZhangMultiTILT:2011} applied
TILT to rectify texts in natural scenes to improve text
recognition on mobile phones. Zhao \etal~\citep{Quan:2011})
proposed a method for detecting translational symmetry using TILT.
Zhang \etal~\citep{ZhangICCV:11} further generalized the
transforms allowed by TILT to polynomially parameterized ones so
that the shape and pose of low rank textures on generalized
cylindrical surfaces can be recovered. Mobahi
\etal~\citep{Mobahi:2011} used the low rank textures recovered by
TILT as the building block for modeling urban scenes and
reconstructing the 3D structure.

TILT was inspired by Robust Alignment by Sparse and Low-rank
decomposition (RASL)~\citep{PengRASL:10}, which has been
successfully applied to align faces and video frames. TILT and
RASL share the same mathematical model, namely after an
appropriate transform a data matrix can be decomposed into a low
rank component and a sparse one (see \eqref{Eq:TILT_1}). The only
difference between TILT and RASL is that the data matrix in TILT
is an image patch, while that in RASL is multiple images or
frames, each being a column of the matrix. So in the sequel, we
just focus on TILT.

The existing most efficient solver for TILT is based on the
alternating direction method
(ADM)~\citep{ZhangTILT:11,LinADM:2010}. It has been adopted by all
the researchers that use TILT for their problems. However, it
still requires multiple seconds to rectify a small sized image
patch, making the applications of TILT far from being interactive.
Another critical drawback of the existing solver is that it was
derived out of intuition by simply mimicing the ADM for the robust
principal component analysis (RPCA) problem presented
in~\citep{LinADM:2010}. In the literature of ADM in the
Gauss-Seidel fashion, almost all the convergence results were
proven under the case of \emph{two} variables, while the inner
loop of TILT problem (see (\ref{Eq:TILT_3})) has \emph{three}
variables. Hence the convergence of ADM for the inner loop of TILT
is not theoretically guaranteed. In our experiments, we did find
many examples that the existing solver for TILT failed to produce
a correct solution (e.g., see Figures~\ref{fig:representative}(c),
\ref{fig:representative}(d), \ref{fig:convergence},
\ref{fig:empirical} and \ref{fig:robustness}). Consequently, to
make the algorithm workable, its parameters have to be carefully
tuned in order to trade off between convergence speed and
robustness at the best (c.f. last paragraph of
Section~\ref{sec:ADM_TILT}), which is difficult for different
applications. The above drawbacks of the existing algorithm
motivated us to design a more efficient algorithm for TILT, with a
theoretical guarantee on the convergence of its inner loop.

We observe that the original method does not always converge to a
correct solution because the inner loop of the original TILT
problem have three variables. This motivates us to cancel one of
the variables and solve an equivalent TILT problem that has only
two variables in its inner loop. Then using the recently developed
linearized alternating direction method with adaptive penalty
(LADMAP)~\citep{LinNIPS:2011} and some good properties of the
annihilation matrix, the inner loop can be solved efficiently and
with a convergence guarantee. We further speed up the computation
by warm starts in two ways. First, we initialize the variables in
the inner loop by their values when they exit the previous inner
loop. This gives the variables very good initial values. So the
number of iterations in the inner loop can be greatly cut. Second,
as singular value decomposition (SVD) is the major bottleneck of
computation, we also solve SVD by warm start, where the singular
vectors and singular values are initialized as those in the
previous iteration. The update of SVD is also based on a recently
developed technique of optimization with orthogonality
constraints~\citep{OptM-Wen-Yin-2010}. As a result, our new
algorithm, called LADMAP with warm starts (LADMAP+WS), can be at
least \emph{five times faster} than the original method and has a
convergence guarantee for the inner loop.

The rest of this paper is organized as follows. In
Section~\ref{sec:tilt} we review the TILT problem and the ADM
method for solving it. In Section~\ref{sec:ladmws} we introduce the
LADMAP method for solving the inner loop of TILT and the variable
warm start technique. In Section~\ref{sec:details}, we present
some important details of applying LADMAP to TILT so that the
computations can be made efficient. In Section~\ref{sec:SVDWS}, we
show the warm start technique to compute SVD. Experimental results
on both simulated and real data are reported in
Section~\ref{sec:exper}. Finally, we conclude our paper in
Section~\ref{sec:concl}.

\section{Transform Invariant Low-rank Textures}\label{sec:tilt}
In this section, we first briefly review the mathematical
model of TILT. Then we introduce its corresponding convex surrogate
and the existing ADM based method for solving it.

\subsection{Mathematical Model}
Consider a 2D texture as a matrix $A\in \mathbb{R}^{m\times n}$.
It is called a low-rank texture if $r \ll \min(m, n)$, where $r =
\mathrm{rank}(A)$. However, a real texture image is hardly an
ideal low-rank texture, mainly due to two factors: 1. it undergoes
a deformation, e.g., a perspective transform from 3D scene to 2D
image; 2. it may be subject to many types of corruption, such as
noise and occlusion.

So if we could rectify a deformed texture $D$ with a proper
inverse transform $\tau$ and then remove the corruptions $E$, then
the resulting texture $A$ will be low rank. This inspires the
following mathematical model for TILT~\citep{ZhangTILT:11}:
\begin{equation}\label{Eq:TILT_1}
    \min_{A,E,\tau} \mathrm{rank}(A)+\lambda\|E\|_0, ~s.t.~D\circ\tau=A+E,
\end{equation}
where $\tau: \mathbb{R}^2\rightarrow \mathbb{R}^2$
belongs to a certain group of transforms, e.g., affine transforms, perspective transforms, and general cylindrical transforms~\citep{ZhangICCV:11}, $\|E\|_0$ denotes the number of non-zero entries in $E$,
and $\lambda > 0$ is a weighting parameter which trades
off the rank of the underlying texture and the sparsity of the corruption.

\subsection{The Existing Method for Solving TILT}\label{sec:ADM_TILT}
The above problem (\ref{Eq:TILT_1}) is not directly tractable,
because the rank of a matrix and the $\ell_0$-norm are discrete
and nonconvex, thus solving the optimization problem is NP-hard.
As a common practice, $\mathrm{rank}$ and $\ell_0$-norm could be
replaced by the nuclear norm~\citep{Candes-2011-RPCA} (the sum of
the singular values) and $\ell_1$-norm~\citep{Donoho2006-CPAM}
(the sum of the absolute values of entries), respectively. This
yields the following optimization problem with a convex objective
function~\citep{ZhangTILT:11}:
\begin{equation}\label{Eq:TILT_2}
    \min_{A,E,\tau} \|A\|_*+\lambda\|E\|_1, ~s.t.~D\circ\tau=A+E.
\end{equation}
The above problem is \emph{not} a convex program as the constraint
is nonconvex. So Zhang \etal~\citep{ZhangTILT:11} proposed to
linearize $D\circ\tau$ at the previous $\tau^{i}$ as
$D\circ(\tau^{i}+\Delta\tau)\approx D\circ\tau^i + J\Delta\tau$
and determine the increment $\Delta\tau$ in the transform by
solving
\begin{equation}\label{Eq:TILT_3}
    \min_{A,E,\Delta\tau} \|A\|_*+\lambda\|E\|_1, ~s.t.~D\circ\tau^i+J\Delta\tau=A+E,
\end{equation}
where $J$ is the the Jacobian: derivative of the image with
respect to the transform parameters. $J$ is full column rank. We
call the iterative procedure to solve (\ref{Eq:TILT_3}) the inner
loop for solving TILT. When the increment $\Delta\tau$ is
computed, the transform is updated as
$$\tau^{i+1}=\tau^i+\Delta\tau.$$
In the following, for brevity when we are focused on the inner
loop, for simplicity we may omit the superscripts $i+1$ or $i$ of
variables. This should not cause ambiguity by referring to the
context.

Zhang \etal~\citep{ZhangTILT:11} proposed an ADM based method to solve (\ref{Eq:TILT_3}). It is to minimize the augmented Lagrangian function
of problem (\ref{Eq:TILT_3}):
\begin{equation*}\label{Eq:Lagrangian}
\begin{array}{rl}
  &L(A,E,\Delta\tau,Y,\mu)\\
  =&\|A\|_*+\lambda\|E\|_1 +\langle Y,D\circ\tau+J\Delta\tau-A-E\rangle\\
  &+ \frac{\mu}{2}\|D\circ\tau+J\Delta\tau-A-E\|_F^2,
\end{array}
\end{equation*}
with respect to the unknown variables \emph{alternately} (hence
the name ADM), where $Y$ is the Lagrange multiplier, $\langle A,B
\rangle\equiv \mbox{tr}(A^TB)$ is the inner product, $\|\cdot\|_F$
is the Frobenius norm, and $\mu>0$ is the penalty parameter. The
ADM in~\citep{ZhangTILT:11} goes as follows: \vspace{6pt}
\begin{eqnarray}
 A_{k+1} &=& \argmin_{A}L(A,E_k,\Delta\tau_k,Y_k,\mu_k), \label{eq:update_A}\\
 E_{k+1} &=& \argmin_{E}L(A_{k+1},E,\Delta\tau_k,Y_k,\mu_k), \label{eq:update_E}\\
 \Delta\tau_{k+1} &=& \argmin_{\Delta\tau}L(A_{k+1},E_{k+1},\Delta\tau,Y_k,\mu_k), \label{eq:update_detla_Tau} \\
 Y_{k+1} &=& Y_k + \mu_k (D\circ\tau+J\Delta\tau_{k+1}-A_{k+1}-E_{k+1}), \nonumber\\
 \mu_{k+1} &=& \rho\mu_k,\nonumber
 \vspace{6pt}
\end{eqnarray}
where $\rho>1$ is a constant. All subproblems (\ref{eq:update_A})-(\ref{eq:update_detla_Tau}) have closed form solutions~\citep{ZhangTILT:11} (cf. (\ref{Eq:A_1}) and (\ref{Eq:E_1})). More complete details of the algorithm
can be found as Algorithm 1 in~\citep{ZhangTILT:11}.

Although the above ADM in the Gauss-Seidel fashion empirically
works well in most cases, there is no theoretical guarantee on its
convergence. There are two factors that may result in its
non-convergence. Namely, all the previous convergence results of
ADM in the Gauss-Seidel fashion were proven under the conditions
that the number of unknown variables are only
\emph{two}\footnote{For the case of more than two variables, ADM
with some appropriate modifications, such as introducing a dummy
variable and treating all the original variables as a super
block~\citep{Bertsekas-Parallel} or introducing a Gaussian back
substitution~\citep{He2011-Gaussian}, can be proven to converge.}
and the penalty parameter is \emph{upper bounded}. In contrast,
the inner loop of TILT has three variables and its penalty
parameter is not upper bounded. As one will see, the naive ADM
algorithm above may not produce a correct solution. In particular,
the choice of $\rho$ is critical to influence the number of
iterations in the inner loop and the quality of solution. If
$\rho$ is small, there will be a lot of iterations but the
solution is very likely to be correct. If $\rho$ is large, the
number of iterations is small but an incorrect solution may be
produced. So one has to tune $\rho$ very carefully so that the
number of iterations and the quality of solution can be best
traded off. This is difficult if the textures are different. So in
this paper we aim at addressing both the computation speed and the
convergence issue of the inner loop in the original method.

\section{Solving the Inner Loop by LADMAP}\label{sec:ladmws}
In this section, we show how to apply a recently developed method
LADMAP to solve the inner loop of TILT. We first reformulate
(\ref{Eq:TILT_3}) so that $\Delta\tau$ is canceled and hence
LADMAP can be applied.

\subsection{Sketch of LADMAP}
LADMAP~\citep{LinNIPS:2011} aims at solving the following
type of convex programs:
\begin{equation}\label{Eq:LADMAP}
    \min_{\x,\y} f(\mathbf{x})+g(\mathbf{y}),
    ~s.t.~\mathcal{A}(\mathbf{x})+\mathcal{B}(\mathbf{y})= \mathbf{c},
\end{equation}
where $\x$, $\y$ and $\bf{c}$ could be either
vectors or matrices, $f$ and $g$ are convex functions,
and $\mathcal{A}$ and $\mathcal{B}$ are linear mappings.

If the original ADM is applied to (\ref{Eq:LADMAP}), $\x$ and $\y$ are updated by minimizing the augmented Lagrangian function of (\ref{Eq:LADMAP}), resulting in the following updating scheme:
\begin{eqnarray}
\mathbf{x}_{k+1}
&=&\arg\min\limits_{\mathbf{x}}f(\mathbf{x})\nonumber\\
&&+ \frac{\mu_k}{2}\|\mathcal{A}(\mathbf{x}) +
\mathcal{B}(\mathbf{y}_{k}) -
\mathbf{c}+\gamma_k/\mu_k\|^2,\label{eq:update_x}\\
\mathbf{y}_{k+1}
&=&\arg\min\limits_{\mathbf{y}}g(\mathbf{y})\nonumber\\
&&+ \frac{\mu_k}{2}\|\mathcal{B}(\mathbf{y})+\mathcal{A}(\mathbf{x}_{k+1}) -
\mathbf{c}+\gamma_k/\mu_k\|^2,\label{eq:update_y}\\
\mathbf{\gamma}_{k+1} &=& \mathbf{\gamma}_k +
\mu_k[\mathcal{A}(\mathbf{x}_{k+1}) +
\mathcal{B}(\mathbf{y}_{k+1})-\mathbf{c}],\label{eq:update_lambda}
\end{eqnarray}
where $\mathbf{\gamma}$ is the Lagrange multiplier and $\mu$ is
the penalty parameter. In many problems, $f$ and $g$ are vector or
matrix norms, and the subproblems (\ref{eq:update_x}) and
(\ref{eq:update_y}) have closed-form solutions when $\mathcal{A}$
and $\mathcal{B}$ are identity
mappings~\citep{CaiSVT:2010,Donoho95-denoising,Liu-2010-LRR,Yang-2010-ADMl1},
hence easily solvable. However, when $\mathcal{A}$ and
$\mathcal{B}$ are not identity mappings, subproblems
(\ref{eq:update_x}) and (\ref{eq:update_y}) may not be easy to
solve. So Lin \etal~\citep{LinNIPS:2011} proposed linearizing the
quadratic penalty term in the augmented Lagrangian function and
adding a proximal term in (\ref{eq:update_x}) and
(\ref{eq:update_y}) for updating $\x$ and $\y$, resulting in the
following updating scheme:
\begin{eqnarray}
\mathbf{x}_{k+1}
&=&\arg\min\limits_{\mathbf{x}}f(\mathbf{x})+\dfrac{\mu_k\eta_A}{2}\|\mathbf{x}-\mathbf{x}_k \nonumber\\
&&\hspace{-3em}+\mathcal{A}^*(\gamma_k+\mu_k(\mathcal{A}(\mathbf{x}_k)
+ \mathcal{B}(\mathbf{y}_{k})-\mathbf{c}))/(\mu_k\eta_A)\|^2,\label{eq:update_x'}\\
\mathbf{y}_{k+1}
&=&\arg\min\limits_{\mathbf{y}}g(\mathbf{y})+\frac{\mu_k\eta_B}{2}\|\mathbf{y}-\mathbf{y}_k \nonumber\\
&&\hspace{-3em}+\mathcal{B}^*(\gamma_k+
\mu_k(\mathcal{A}(\mathbf{x}_{k+1})+
\mathcal{B}(\mathbf{y}_{k})-\mathbf{c}))/(\mu_k\eta_B)\|^2,\label{eq:update_y'}
\end{eqnarray}
and $\lambda$ is still updated as (\ref{eq:update_lambda}), where
$\eta_A>0$ and $\eta_B>0$ are some parameters. Then one can see
the new subproblems (\ref{eq:update_x'}) and (\ref{eq:update_y'})
can have closed-form solutions again when $f$ and $g$ are norms.
Lin \etal also proposed a strategy to adaptively update the
penalty parameter $\mu$ \footnote{Please refer to (\ref{Eq:mu})
and (\ref{Eq:rho}). For succinctness, we do not repeat it here.
Disregarding the adaptive penalty, LADMAP is a special case of the
generalized ADM~\citep{Deng12-global} and is closely related to
the predictor corrector proximal multiplier
method~\citep{Chen94-proximal} which updates variables parallelly
rather than alternately.} and proved that when $\mu$ is
non-decreasing and upper bounded, and $\eta_A>\|\mathcal{A}\|^2$
and $\eta_B>\|\mathcal{B}\|^2$, $(\mathbf{x}_k, \mathbf{y}_k)$
converges to an optimal solution to (\ref{Eq:LADMAP}), where
$\|\mathcal{A}\|$ and $\|\mathcal{B}\|$ are the operator norms of
$\mathcal{A}$ and $\mathcal{B}$, respectively. For more details,
please refer to \citep{LinNIPS:2011}.

\subsection{Reformulating the Inner Loop of TILT}\label{sec:reformulate}
As almost all existing convergence results of ADM or linearized
ADM in the Gauss-Seidel fashion are proven in the case of two
variables, while the inner loop of TILT has three variables, we
aim at canceling one variable by taking advantage of the special
structure of the problem.

We notice that $\Delta\tau$ does not appear in the objective function of (\ref{Eq:TILT_3}) and it is easy to see that if $(A^*,E^*)$ is an optimal solution, then the optimal $\Delta\tau$ must be
\begin{equation}\label{Eq:Delta_tau}
    \Delta\tau^* = (J^TJ)^{-1}J^T(A^*+E^*-D\circ\tau),
\end{equation}
as $(A^*,E^*,\Delta\tau^*)$ must satisfy the linear constraint in
(\ref{Eq:TILT_3}). So we have to find an optimization problem to
obtain $(A^*,E^*)$.

By plugging (\ref{Eq:Delta_tau}) into the linear constraint in
(\ref{Eq:TILT_3}), one can see that $(A^*,E^*)$ must satisfy the
following constraint:
\begin{equation}\label{Eq:Delta_tau_canceled1}
J^\perp A+J^\perp E=J^\perp (D\circ\tau),
\end{equation}
where $J^\perp = I - J(J^TJ)^{-1}J^T$ and $I$ is the identity
matrix. So we have an optimization problem for $(A^*,E^*)$:
\begin{equation}\label{Eq:TILT_LADMAP}
    \min_{A,E} \|A\|_*+\lambda\|E\|_1,
    ~s.t.~J^\perp A+J^\perp E=J^\perp (D\circ\tau).
\end{equation}

We can prove the following proposition.
\begin{proposition}\label{pro:1}
Problems (\ref{Eq:TILT_3}) and  (\ref{Eq:TILT_LADMAP})
have the same optimal solution $(A^*, E^*)$.
\end{proposition}
Proof. We only have to prove that the constraints for $(A, E)$ do not change. The constraints for $(A, E)$ in (\ref{Eq:TILT_3}) and (\ref{Eq:TILT_LADMAP}) can be written as
\begin{equation*}\label{Eq:constraint_set1}
S_1 = \{(A,E)|\exists \Delta\tau \mbox{ such that } D\circ\tau + J\Delta\tau=A+E\},
\end{equation*}
and
\begin{equation*}\label{Eq:constraint_set2}
S_2 = \{(A,E)| J^\perp A+J^\perp E=J^\perp (D\circ\tau)\},
\end{equation*}
respectively.
If $(A, E)\in S_1$, then we can multiply $J^\perp$ to both sides of
$D\circ\tau + J\Delta\tau=A+E$ to see that $(A, E)\in S_2$.
If $(A, E)\in S_2$,
then $D\circ\tau-A-E \in \mathrm{null}(J^\perp)$.
Since $\mathrm{null}(J^\perp)=\mathrm{span}(J)$,
there exists $\Delta\tau$ such that $D\circ\tau-A-E=J\Delta\tau$.
Thus $(A, E)\in S_1$.
$\Box$

One can easily check that $J^\perp$ has the following nice
properties:
\begin{equation}\label{Eq:J_properties}
    (J^\perp)^T=J^\perp
~~\mathrm{and}
    ~~J^\perp J^\perp = J^\perp.
\end{equation}
Moreover, denote the operator norm of $J^\perp$ as $\|J^\perp\|$. Then we have
\begin{proposition}\label{pro:2}
$\|J^\perp\| = 1$.
\end{proposition}
Proof: From (\ref{Eq:J_properties})
we have $(J^\perp)^2 = J^\perp$. So the eigenvalues $\lambda$
of $J^\perp$ satisfies $\lambda^2 - \lambda = 0$.
Thus, the eigenvalues of $J^\perp$ are either 1 or 0.
As $J^\perp$ is symmetric, $\|J^\perp\| = 1$.
$\Box$

Applying LADMAP ((\ref{eq:update_x'}),(\ref{eq:update_y'}), and
(\ref{eq:update_lambda})) directly to (\ref{Eq:TILT_LADMAP})
(where $\mathbf{x}$, $\mathbf{y}$, and $\mathbf{\gamma}$ are $A$,
$E$, and $Y$, respectively), with some algebra we have the
following updating scheme:
\begin{eqnarray}
  A_{k+1} &=& \argmin_{A} \|A\|_* + \dfrac{\mu_k}{2}\| A-M_k\|_F^2, \label{Eq:A}\\
  E_{k+1} &=& \argmin_{E} \lambda\|E\|_1 + \dfrac{\mu_k}{2}\| E-N_k\|_F^2, \label{Eq:E}\\
  Y_{k+1} &=& Y_k + \mu_kJ^\perp(A_{k+1}+E_{k+1}-D\circ\tau),\label{Eq:Y}\\
  \mu_{k+1} &=& \min(\mu_{max}, \rho\cdot \mu_k), \label{Eq:mu}
\end{eqnarray}
where
\begin{eqnarray}
M_k&=&A_k-J^\perp(A_k+E_k-D\circ \tau + Y_k/\mu_k),\\
N_k&=&E_k - J^\perp(A_{k+1}+E_k-D\circ \tau + Y_k/\mu_k),
\end{eqnarray}
$\mu_{max}$ is an upper bound of ${\mu_k}$ and
\begin{equation}\label{Eq:rho}
\rho=  \left\{
      \begin{array}{ll}
        \rho_0, & \hbox{if }\mu_k\cdot \max(\|A_{k+1}-A_k\|_F,\\
        & \hspace{1em} \|E_{k+1}-E_k\|_F)/\|J^\perp(D\circ\tau)\|_F<\varepsilon_2, \\
        1, & \hbox{otherwise,}
      \end{array}
    \right.
\end{equation}
in which $\rho_0 \geq 1$ is a constant and $\varepsilon_2 > 0$ is
a small threshold. Note that in (\ref{Eq:A})-(\ref{Eq:E}) we have
utilized the properties of $J^\perp$ in (\ref{Eq:J_properties})
and $\|J^\perp\| = 1$. The updating scheme (\ref{Eq:mu}) and
(\ref{Eq:rho}) for $\mu$ comes from the adaptive penalty strategy
in LADMAP~\citep{LinNIPS:2011}.

The closed form solution to (\ref{Eq:A}) is as follows
\citep{CaiSVT:2010}:
\begin{equation}\label{Eq:A_1}
    A_{k+1} = S_{\frac{1}{\mu_k}}(M_k),
\end{equation}
where $S(\cdot)$ is the singular value shrinkage operator:
\begin{equation}\label{Eq:S_operator}
    S_\varepsilon(W) = UT_\varepsilon(\Sigma) V^T,
\end{equation}
in which $U\Sigma V^T$ is the SVD of $W$ and
$T_\varepsilon(\cdot)$ is the scalar shrinkage operator defined as
\citep{Donoho95-denoising}:
$$T_\varepsilon(x)=\sgn(x)\max(|x|-\varepsilon,0).$$

Subproblem (\ref{Eq:E}) also has a closed form solution as follows
\citep{Yang-2010-ADMl1}:
\begin{equation}\label{Eq:E_1}
    E_{k+1} = T_{\frac{\lambda}{\mu_k}}(N_k).
\end{equation}
The iterations (\ref{Eq:A})-(\ref{Eq:mu}) stop when the following two conditions are satisfied:
    \begin{equation}\label{Eq:Cond1}
      \|J^\perp(A_{k+1}+E_{k+1}-D\circ\tau)\|_F/\|J^\perp(D\circ\tau)\|_F< \varepsilon_1
    \end{equation}
and
  \begin{equation}\label{Eq:Cond2}
      \mu_k\cdot \max(\|A_{k+1}-A_k\|_F,\|E_{k+1}-E_k\|_F)/\|
      J^\perp(D\circ\tau)\|_F < \varepsilon_2.
  \end{equation}

These two stopping criteria are set
based on the KKT conditions of problem (\ref{Eq:TILT_LADMAP}).
Details of the deduction can be found in \citep{LinNIPS:2011}.

\subsection{Warm Starting Variables in the Inner Loop}
Since the number of iterations in the inner
loop greatly affects the efficiency of the whole algorithm,
we should provide good initial values to the variables in the inner loop so that the number of iterations can be reduced.

The original algorithm initializes $A$ as the input $D\circ\tau$
and $E$ and $Y$ as $0$. Such a cold start strategy does not
utilize any information from the previous loop. We observe that
solutions from the previous inner loop are good initializations
for next inner loop, because the difference in $D\circ\tau$ in
successive inner loops becomes smaller and smaller. So their
solutions should be close to each other. This motivates us to use
the following warm start strategy for the variables:
\begin{equation}\label{Eq:warm_start}
    A_{0}^{i+1} = A_{\infty}^i,~~E_{0}^{i+1} = E_{\infty}^i,~~\mathrm{and}~~Y_{0}^{i+1} = Y_{\infty}^i,
\end{equation}
where the subscripts and superscripts are indices of the inner and outer loops, respectively.
The subscripts $0$ and $\infty$ stand for the initial and final values in an inner loop, respectively.

We summarize our LADMAP with variable warm start (LADMAP+VWS)
approach for solving TILT in Algorithm~\ref{alg:LADMWS}. Note that
we only change the part of the inner loop. The rest of the
algorithm is inherited from that in~\citep{ZhangTILT:11}.

\begin{algorithm}[t]
\caption{{\bf (TILT via LADMAP+VWS)}}\label{alg:LADMWS}
\begin{algorithmic}
\STATE \textbf{Input:} A rectangular window $D \in
\mathbb{R}^{m\times n}$ in an image, the initial parameters
$\tau^0$ of the transform, weight parameter $\lambda > 0$, $\rho_0
\geq 1$ and $\mu_{max} > 0$. \STATE \textbf{Initialize:} $A^0=
D\circ\tau^0$, $E^0=0$, $Y^0=0$, and $i=0$. \STATE \textbf{While}
not converged \textbf{Do} \STATE $\qquad$ \textbf{Step 1:}
normalize the image and compute the Jacobian w.r.t. the
transform:\vspace{-2mm}
\begin{equation*}
D\circ\tau^{i} \leftarrow
\frac{D\circ\tau^{i}}{\|D\circ\tau^{i}\|_F}, \quad J^{i+1}
\leftarrow \frac{\partial}{\partial
\zeta}\left(\frac{D\circ\zeta}{\|D\circ\zeta\|_F}\right)
\Bigr|_{\zeta=\tau^{i}};\vspace{-2mm}
\end{equation*}
\STATE  $\qquad$ \textbf{Step 2:} solve the linearized convex optimization:
\begin{equation*}
\min_{A,E} \; \|A\|_* + \lambda\|E\|_1, \quad \mathrm{s.t.} \quad
(J^{i+1})^\perp(D\circ\tau^i) = (J^{i+1})^\perp(A + E),
\end{equation*}
\qquad with the initial conditions:
$ A_{0}^{i+1} = A_{\infty}^{i}$, $E_{0}^{i+1} = E_{\infty}^{i}$
and $Y_{0}^{i+1} = Y_{\infty}^{i}$, $\mu_0 > 0$, $k = 0$:\vspace{+2mm}
\STATE  $\qquad$\textbf{While} (\ref{Eq:Cond1}) or (\ref{Eq:Cond2}) are not satisfied \textbf{Do}\vspace{-3mm}
\begin{equation*}
\begin{array}{l}
   A^{i+1}_{k+1} = \\
   \quad S_{\frac{1}{\mu_k}}\Big( A^{i+1}_k-(J^{i+1})^\perp(A^{i+1}_k+E^{i+1}_k-D\circ\tau^i+\mu_k^{-1}Y^{i+1}_k) \Big),\\
   E^{i+1}_{k+1} = \\
   \quad T_{\frac{\lambda}{\mu_k}}\Big(E^{i+1}_k-(J^{i+1})^\perp(A^{i+1}_{k+1}+E^{i+1}_k-D\circ\tau+\mu_k^{-1}Y^{i+1}_k)\Big),\\
   Y^{i+1}_{k+1} = Y^{i+1}_k - \mu_k (J^{i+1})^\perp(A^{i+1}_{k+1}+E^{i+1}_{k+1}-D\circ\tau^i),\\
   \mu_{k+1} = \min(\mu_{max}, \rho\cdot \mu_k),\mbox{ where $\rho$ is given by }(\ref{Eq:rho}),\\
   k\leftarrow k+1.\vspace{-3mm}
\end{array}
\end{equation*}
\STATE  $\qquad$ \textbf{End While}\vspace{+2mm} \STATE  $\qquad$
\textbf{Step 3:} compute optimal $\Delta\tau^*$ using Eq.
(\ref{Eq:Delta_tau}), where $A^*=A^{i+1}_\infty$,
$E^*=E^{i+1}_\infty$;\vspace{+2mm} \STATE  $\qquad$ \textbf{Step
4:} update transform: $\tau^{i+1} = \tau^i + \Delta\tau^*$, $i
\leftarrow i+1$; \vspace{+1mm} \STATE \textbf{End While} \STATE
\textbf{Output:} converged solution $(A^*, E^*, \tau^*)$.
\end{algorithmic}
\label{alg:tilt}
\end{algorithm}

\section{Implementation Details}\label{sec:details}
In the previous section we introduce the basic ideas of LADMAP with
variable warm start (LADMAP+VWS) algorithm for solving the inner
loop of TILT (\ref{Eq:TILT_LADMAP}). However, there are still some
details that need to be handled carefully so that the computation
can be the most efficient. In this section, we first show how to
compute with $J^\perp$ efficiently, then we discuss how to modify
the algorithm in order to accommodate some additional constraints
when putting TILT to real applications.

\subsection{Multiplying with $J^\perp$}
In Algorithm \ref{alg:LADMWS}, $J$ is actually an order-3
tensor~\citep{ZhangTILT:11}. When computing $J$ is actually
arranged in an $mn\times p$ matrix, where $m\times n$ is the size
of $D\circ\tau$ and $p$ is the number of variables to parameterize
the transform $\tau$. When multiplying $J^\perp$ with a matrix $M$
we actually mean to rearrange the matrix $M$ into an $mn\time 1$
vector and then multiply it with $J^\perp$. However, $J^\perp$ is
an $mn\times mn$ matrix, which is huge and often cannot be fit
into the memory. Moreover, the computation cost will be
$O((mn)^2)$, which is also unaffordable. So we cannot form
$J^\perp$ explicitly and then multiply it with a vectorized
matrix. Recall that $J^\perp=I-J(J^TJ)^{-1}J^T$, we may overcome
this difficulty by multiplying successively:
\begin{equation}\label{Eq:Jperpv}
    J^\perp v = v - J\cdot((J^TJ)^{-1}\cdot(J^T v)),
\end{equation}
whose computation cost is only $O(pmn)$. Note that $(J^TJ)^{-1}$ is only a $p\times p$
small matrix which can be pre-computed and saved. So the above strategy is much more
efficient in both memory and computation.

\subsection{Initializing $Y$}
In Algorithm \ref{alg:LADMWS}, we have to multiply
$(J^{i+1})^\perp$ with three different matrices. If we initialize
$Y^{i+1}_0$ in the subspace spanned by $(J^{i+1})^\perp$, then
$Y^{i+1}_k$ is always in the subspace spanned by $(J^{i+1})^\perp$
during the iteration. Then thanks to the idempotency of
$(J^{i+1})^\perp$, we have $(J^{i+1})^\perp Y^{i+1}_k=Y^{i+1}_k$
and hence $A$ and $E$ can be updated as
\begin{small}
\begin{equation*}
\begin{array}{l}
   A^{i+1}_{k+1} = \\
   \quad S_{\frac{1}{\mu_k}}\Big( A^{i+1}_k-(J^{i+1})^\perp(A^{i+1}_k+E^{i+1}_k-D\circ\tau^i)-\mu_k^{-1}Y^{i+1}_k \Big),\\
   E^{i+1}_{k+1} = \\
   \quad T_{\frac{\lambda}{\mu_k}}\Big(E^{i+1}_k-(J^{i+1})^\perp(A^{i+1}_{k+1}+E^{i+1}_k-D\circ\tau^i)-\mu_k^{-1}Y^{i+1}_k\Big).
\end{array}
\end{equation*}
\end{small}
Then one can see that we only have to multiply
$(J^{i+1})^\perp$ with two different matrices,
$A^{i+1}_k+E^{i+1}_k-D\circ\tau^i$ and
$A^{i+1}_{k+1}+E^{i+1}_k-D\circ\tau^i$, because
$(J^{i+1})^\perp(A^{i+1}_{k+1}+E^{i+1}_{k+1}-D\circ\tau^i)$ is
used for updating both $Y^{i+1}_{k+1}$ and $A^{i+1}_{k+2}$. This
saves one multiplication of multiplying with $(J^{i+1})^\perp$ for
each iteration.

To combine with the warm start technique, $Y$ should be initialized as
\begin{equation}\label{Eq:warm_start_Y}
Y_{0}^{i+1} = (J^{i+1})^\perp Y_{\infty}^i.
\end{equation}
Because the dual problem of the inner loop (\ref{Eq:TILT_LADMAP}) is
\begin{equation*}\label{Eq:dual_TILT}
\max_{Y} \langle J^\perp Y, D\circ \tau\rangle ~s.t.~\|J^\perp
Y\|_2\leq 1, \|J^\perp Y\|_\infty \leq \lambda^{-1},
\end{equation*}
where $\|\cdot\|_\infty$ is the maximum absolute value in a
matrix, we see that the optimal $Y$ can be chosen in
$\mbox{span}(J^\perp)$. So constraining $Y$ in
$\mbox{span}(J^\perp)$ does not affect the convergence of
LADMAP+VWS to the optimal solution. As
$Y_{\infty}^i\in\mbox{span}((J^{i})^\perp)$, when $J^{i+1}$ is
close to $J^i$, $Y_0^{i+1}\approx Y_{\infty}^i$. So
(\ref{Eq:warm_start_Y}) makes good combination of warm starting
$Y$ and making $Y\in\mbox{span}(J^\perp)$.

\subsection{Handling Additional Constraints on $\tau$}
As is discussed in \citep{ZhangTILT:11}, additional constraints
should be imposed on $\tau$ so as to eliminate degenerate or
trivial solutions, e.g., $\tau^*$ being 0. Typical constrains
include that both the center and area of the rectangle being
fixed. These additional constraints can be formulated as $l$
linear constraints on $\Delta\tau$ \citep{ZhangTILT:11}:
\begin{equation}\label{Eq:add_constraints}
    Q\cdot\Delta\tau = 0,
\end{equation}
where $Q\in \mathbb{R}^{l\times p}$.

Following the same idea as that in Section~\ref{sec:reformulate},
we aim at eliminating $\Delta\tau$ with the $l$ additional
constraints. As $\Delta\tau$ needs to satisfy both the linear
constraints in Eq. (\ref{Eq:add_constraints}) and that in problem
(\ref{Eq:TILT_3}), the overall constraints for $\Delta\tau$ are
\begin{equation}\label{Eq:two_linear_constraint}
\left[
\begin{array}{c}
           J \\
           Q \\
         \end{array}
           \right]
\Delta\tau ~~= \left[
         \begin{array}{c}
           A+E-D\circ\tau\\
           0\\
         \end{array}
       \right].
\end{equation}
Similar to the deductions in Section~\ref{sec:reformulate}, we can
have an equivalent problem:
\begin{equation}\label{Eq:reformulated_TILT2}
    \min_{A,E} \|A\|_*+\lambda\|E\|_1, ~s.t.~W A+W
    E=W(D\circ\tau),
\end{equation}
where
\begin{equation*}\label{Eq:W}
     W = \left[
\begin{array}{c}
         I - J(J^TJ+Q^TQ)^{-1}J^T \\
           -Q(J^TJ+Q^TQ)^{-1}J^T\\
         \end{array}
           \right].
\end{equation*}

Matrix $W$ also enjoys a nice property similar to $J^\perp$:
\begin{equation}\label{Eq:WTW_property}
   W^TW = I-J(J^TJ+Q^TQ)^{-1}J^T,
\end{equation}
which can be utilized to reduce the computational cost. Moreover,
$\|W\|_2=1$. Then with some algebra LADMAP applied to the above
problem goes as follows:
\begin{equation*}\label{Eq:alg_new_constraints1}
\begin{array}{rcl}
  A_{k+1} &=& S_{\frac{1}{\mu_k}}(M_{k}), \\
  E_{k+1} &=& T_{\frac{\lambda}{\mu_k}}(N_k), \\
  Y_{k+1} &=& Y_k + \mu_k\cdot W(A_{k+1}+E_{k+1}-D\circ\tau), \\
  \mu_{k+1} &=& \min\big(\rho\cdot\mu_k, \mu_{max} \big),
\end{array}
\end{equation*}
where
\begin{equation*}\label{Eq:newvars1}
\begin{array}{rcl}
  M_{k} &=& A_{k}- \mu_k^{-1}W^TY_k - W^TW(A_k+E_k-D\circ\tau), \\
  N_{k} &=& A_{k+1}- \mu_k^{-1}W^TY_k - W^TW(A_{k+1}+E_k-D\circ\tau),
\end{array}
\end{equation*}
and $\rho$ is still computed as (\ref{Eq:rho}).

Thanks to (\ref{Eq:WTW_property}), the multiplication of $W^TW$
with a vectorized matrix can be done similarly as
(\ref{Eq:Jperpv}). To further reduce the computational cost, we
introduce $\tilde{Y}_k=W^T Y_k$ and initialize it in
$\mbox{span}(W^TW)$. Then $M_k$, $N_k$ and $\tilde{Y}_k$ are
computed as follows:
\begin{equation}\label{Eq:newvars2}
\begin{array}{rcl}
  M_{k} &=& A_{k}- \mu_k^{-1}\tilde{Y}_k - W^TW(A_k+E_k-D\circ\tau), \\
  N_{k} &=& A_{k+1}- \mu_k^{-1}\tilde{Y}_k - W^TW(A_{k+1}+E_k-D\circ\tau),\\
  \tilde{Y}_{k+1} &=& \tilde{Y}_k + \mu_k\cdot W^TW(A_{k+1}+E_{k+1}-D\circ\tau).
\end{array}
\end{equation}
Again, $W^TW(A_{k+1}+E_{k+1}-D\circ\tau)$ is used to update both
$M_{k+1}$ and $\tilde{Y}_{k+1}$. In this way, the iterations for
the inner loop can be computed very efficiently.

Now the warm start of $Y$ is replaced by that of $\tilde{Y}$, which is:
\begin{equation}\label{Eq:Y_new_warmstart}
    \tilde{Y}^{i+1}_0 = (W^{i+1})^TW^{i+1} \tilde{Y}^{i}_{\infty}.
\end{equation}
As $l\ll mn$, $(W^{i+1})^TW^{i+1}$ is actually very close to
$(J^{i+1})^\perp$. So (\ref{Eq:Y_new_warmstart}) is both a good
combination of warm starting $\tilde{Y}$ and making
$\tilde{Y}\in\mbox{span}(W^TW)$.
\section{Warm Starting for SVD in the Inner Loop}\label{sec:SVDWS}
As shown in (\ref{Eq:A_1})-(\ref{Eq:S_operator}), to update $A$ we
have to compute the SVD of $M_k$. Unlike other low-rank recovery
problems~\citep{CaiSVT:2010,LinADM:2010,TokAPG:2009}, partial SVD
cannot be used here. This is because partial SVD is faster than
full SVD only when the rank of $A_{k+1}$ is less than
$\min(m,n)/5$~\citep{LinADM:2010}, while when rectifying general
textures this condition is often violated. So computing the full
SVD of $M_k$ is very costly as its complexity is $O(mn\min(m,n))$.
Without exaggeration, the efficiency of computing the full SVD
dominates the computing time of solving TILT. So we have to reduce
the computation cost on full SVD as well.

We observe that, except for the first several steps in the inner
loop, the change in $M_k$ between two iterations is relatively
small. So we may expect that the SVDs of $M_k$ and $M_{k-1}$ in
two successive iterations may be close to each other. This
naturally inspires us to utilize the SVD of $M_{k-1}$ to estimate
the SVD of $M_k$.

To do so, we first formulate the SVD problem as follows:
\begin{equation}\label{Eq:SVD}
\begin{array}{c}
  (U^*,\Sigma^*,V^*)=\argmin_{U,\Sigma,V} F(U,\Sigma,V), \\
  \\
  s.t.~~U^TU=I, ~~\Sigma~\mbox{is diagonal}, ~~\mathrm{and}~~V^TV=I,
\end{array}
\end{equation}
where $F(U,\Sigma,V)=\dfrac{1}{2}\|M-U\Sigma V^T\|_F^2$ and $M\in
\mathbb{R}^{m\times n}$ is the matrix to be decomposed. Without
loss of generality we may assume $m\geq n$. $U\in
\mathbb{R}^{m\times n}$ and $V \in \mathbb{R}^{n\times n}$ are
columnly orthonormal and orthogonal matrices, respectively. As we
can negate the columns of $U$ or $V$, we need not require the
diagonal entries of $\Sigma$ to be nonnegative.

\subsection{Optimization with Orthogonality Constraints}
The usual method to solve a general orthogonality constrained
optimization problem is to search along the geodesic of the
Stiefel manifold \footnote{A Stiefel manifold is a set of columnly
orthonormal matrices whose geodesic distance is induced from the
Euclidian distance of its ambient space.} along the direction of
the gradient of the objective function projected onto the tangent
plane of the manifold~\citep{Edelman-1999-orthogonal}. This may
require SVDs in order to generate feasible points on the geodesic.
Fortunately, Wen and Yin~\citep{OptM-Wen-Yin-2010} recently
developed a technique that does not rely on SVDs, making our warm
start for SVD possible.

Denote the unknown variable as $X$. Suppose the gradient of the
objective function at $X$ is $G$, then the projection of $G$ onto
the tangent plane of the Stiefel manifold at $X$ is
$P=GX^T-XG^T$~\citep{Edelman-1999-orthogonal,OptM-Wen-Yin-2010}.
Instead of parameterizing the geodesic of the Stiefel manifold
along direction $P$ using the exponential function, Wen and
Yin~\citep{OptM-Wen-Yin-2010} proposed generating feasible points
by the Cayley transform:
\begin{equation*}\label{Eq:Caylay_Eg}
    X(\tau) = C(\tau)X,~~\mathrm{where}~~C(\tau)=\left(I+\frac{\tau}{2}P\right)^{-1}\left(I-\frac{\tau}{2}P\right).
\end{equation*}
It can be verified that $X(\tau)$ has the following properties:
\begin{itemize}
\item[1.] $X(\tau)$ is smooth in $\tau$; \item[2.]
$\big(X(\tau)\big)^T X(\tau) = I$, $\forall \tau\in \mathbb{R}$,
given $X^TX=I$; \item[3.] $X(0)=X$; \item[4.]
$\frac{d}{d\tau}X(0)=-P$.
\end{itemize}
So when $\tau>0$ is sufficiently small, $X(\tau)$ can result in a
smaller objective function value than $X$.

$X(\tau)$ could be viewed as reparameterizing the geodesic with
$\tau$, which does not exactly equal to the geodesic distance.
However, when $\tau$ is small it is very close to the geodesic
distance as it is the length of the two segments enclosing the
geodesic~\citep{OptM-Wen-Yin-2010}. When computing $X(\tau)$, no
SVD is required. A matrix inversion and some matrix
multiplications are required instead, which is of much lower cost
than SVD. However, as both matrix multiplication/inversion and SVD
are of the same order of complexity, we have to control the number
of matrix multiplications and inversions, so that our warm start
based method can be faster than directly computing the full SVD.

\subsection{SVD with Warm Start}
Now for our SVD problem (\ref{Eq:SVD}), we can compute the
gradient $(G_U,G_\Sigma,G_V)$ of the objective function $F(U,\Sigma,V)$ with
respect to $(U,\Sigma,V)$ and search on a geodesic on the constraint
manifold $C_U\times C_\Sigma\times C_V$ in the gradient direction for
the next best solution, where $C_U$ is the Stiefel manifold of all
columnly orthonormal $m\times n$ matrices, $C_\Sigma$ is the subspace
of all $n\times n$ diagonal matrices, and $C_V$ is the Stiefel
manifold of all $n\times n$ orthogonal matrices.

We search on the constraint manifold on the following curve:
\begin{equation}\label{Eq:curve_to_search}
     \begin{array}{rcl}
       \vspace{+2mm}
       U(t)&=&\left(I+\frac{t}{2}P_U\right)^{-1}\left(I-\frac{t}{2}P_U\right)U,\\
       \vspace{+2mm}
       \Sigma(t)&=& \Sigma - t\cdot P_\Sigma,\\
       \vspace{+2mm}
       V(t)&=&\left(I+\frac{t}{2}P_V\right)^{-1}\left(I-\frac{t}{2}P_V\right)V,
     \end{array}
\end{equation}
where $P_U=G_U U^T-U G_U^T$ and $P_V=G_VV^T-VG_V^T$ are the
projection of $G_U$ and $G_V$ onto the tangent planes of $C_U$ and
$C_V$ at $U$ and $V$, respectively, and $P_\Sigma=\diag(G_\Sigma)$ is the
projection of $G_\Sigma$ onto $C_\Sigma$. The details of computing the
gradients can be found in Appendix.

Then we may find the optimal $t^*$ such that
\begin{equation}\label{Eq:solve_t}
    t^* = \argmin_t f(t) = \frac{1}{2}\|M-U(t)\cdot \Sigma(t)\cdot V(t)^T\|_F^2.
\end{equation}
As $f(t)$ is a complicated function of $t$, the optimal $t^*$ has to be found by iteration.
This will be costly as many matrix inversions and multiplication will be required. So
we choose to approximate $f(t)$ by a quadratic function via Taylor expansion:
\begin{equation}\label{Eq:ft}
    f(t) \approx f(0)+ f'(0)\cdot t + \frac{1}{2}f''(0)\cdot t^2,
\end{equation}
where $f'(0)$ and $f''(0)$ are the first and second order
derivatives of $f(t)$ evaluated at $0$, respectively. These two
derivatives can be computed efficiently. Details of the deductions
can be found in Appendix. Then we can obtain an approximated
optimal solution $\tilde{t}^*=-f'(0)/f''(0)$ and approximate the
SVD of $M$ as $U(\tilde{t}^*)\Sigma(\tilde{t}^*)V(\tilde{t}^*)^T$.

The warm start SVD method is summarized in Algorithm
\ref{alg:SVDWS}. It is called only when the difference between
$M_k$ and $M_{k-1}$ is smaller than a pre-defined threshold
$\varepsilon_{svd}$.

\begin{algorithm}[tb]
\caption{{\bf (SVD with Warm Start)}}\label{alg:SVDWS}
\begin{algorithmic}
\STATE \textbf{Input}: The decomposed matrices $U_{k-1}$,
$\Sigma_{k-1}$ and $V_{k-1}$ of the SVD of $M_{k-1}$, and a matrix
$M_k$. \STATE \textbf{If} $\|M_{k}-M_{k-1}\|<\varepsilon_{svd}$
(\textbf{Else} do full svd) \STATE  \qquad \textbf{Step 0:}
$M=M_k$, $U(0)=U_{k-1}$, $\Sigma(0)=\Sigma_{k-1}$, and
$V(0)=V_{k-1}$. \STATE  \qquad \textbf{Step 1:} compute the
projected gradients $P_U$, $P_\Sigma$, and $P_V$  of $F$ at
$U(0)$, $\Sigma(0)$, and $V(0)$ using \eqref{Eq:gradient1} to \eqref{Eq:gradient3}, respectively.
\STATE \qquad
\textbf{Step 2:} compute $f'(0)$, $f''(0)$, and
$\tilde{t}^*=-f'(0)/f''(0)$ using \eqref{Eq:projected_gradient1} to \eqref{Eq:projected_gradient3}.
\STATE \qquad \textbf{Step 3:}
compute $U_{k}=U(\tilde{t}^*)$, $\Sigma_{k}=\Sigma(\tilde{t}^*)$,
$V_{k}=V(\tilde{t}^*)$ using (\ref{Eq:curve_to_search}). \STATE
\textbf{Output}: $U_{k}$, $\Sigma_{k}$, and $V_{k}$ as the
decomposed matrices in the SVD of $M_k$.
\end{algorithmic}
\end{algorithm}

Although $U(\tilde{t}^*)\Sigma(\tilde{t}^*)V(\tilde{t}^*)^T$ is an
approximate SVD of $M$, our final goal is to compute the singular
value shrinkage of $M_k$ in order to update $A_{k+1}$ (see
(\ref{Eq:A}), (\ref{Eq:A_1}) and (\ref{Eq:S_operator})), not the
SVD of $M_k$ itself. We can show that when $M_k$ is close enough
to $M_{k-1}$, computing the SVD of $M_k$ approximately still
produces a highly accurate $A_{k+1}$. The corner stone of our
proof is the following pseudocontraction property of the singular
value shrinkage operator:
\begin{eqnarray*}
\begin{array}{l}
   \|S_{\varepsilon}(W_1) - S_{\varepsilon}(W_2)\|_F^2 \leq
\|W_1 - W_2\|_F^2 \\
\quad\quad - \|[S_{\varepsilon}(W_1)-W_1] -
[S_{\varepsilon}(W_2)-W_2]\|_F^2,
\end{array}
\end{eqnarray*}
thanks to Lemma 3.3 of \citep{Pierra:1989} and the fact that
$S_{\varepsilon}(\cdot)$ is the proximal mapping of the nuclear
norm~\citep{CaiSVT:2010}. Then we have:
\begin{equation*}
\begin{array}{l}
   \|S_{\varepsilon}\big(U(\tilde{t}^*)\Sigma(\tilde{t}^*)V(\tilde{t}^*)^T\big)
-S_{\varepsilon}\big(M_{k}\big)\|_F \\
\leq \|U(\tilde{t}^*)\Sigma(\tilde{t}^*)V(\tilde{t}^*)^T - M_{k}\|_F\\
\leq \|U(0)\Sigma(0)V(0)^T - M_{k}\|_F = \|M_{k-1} - M_{k}\|_F.
\end{array}
\end{equation*}
Since we switch to our warm start technique for SVD in the inner
loop only when $M_k$ is very close to $M_{k-1}$ (i.e., $\|M_{k-1}
- M_{k}\| < \varepsilon_{svd}$), it is guaranteed that:
$$\|S_{\varepsilon}\big(U(\tilde{t}^*)\Sigma(\tilde{t}^*)V(\tilde{t}^*)^T\big)
-S_{\varepsilon}\big(M_{k}\big)\|_F < \varepsilon_{svd}.$$ Hence
our approximate SVD for $M_k$ still results in a highly accurate
$A_{k+1}$.

\begin{table*}[t]
\begin{center}
\caption{\textbf{Comparison of the efficiency of algorithms on the
inner loop of TILT only}. The time costs, numbers of iterations,
and optimal objective function values are averaged over 10 trials
on random data.} \label{tab:numerical} {\small
\begin{tabular}{|c||c|c|c|c|c|c|c|c|c|}
\hline \textbf{Size of $D\circ\tau$}
 & \multicolumn{3}{|c|}{~~ADM~~}
 & \multicolumn{3}{|c|}{~LADMAP~}
 & \multicolumn{3}{|c|}{LADMAP+SVDWS}\\
\hline \hline
 & time (s) & \#iter. & obj. func.
 & time (s) & \#iter. & obj. func.
 & time (s) & \#iter. & obj. func.\\
\hline$10\times 10$ & 1.98E-05 & ~~49.2~~ & 6.4836&
\textbf{7.49E-06} &~~20.4~~ & 6.4814&
8.76E-06 & ~~20.9~~ & 6.4815\\
\hline $50\times 50$ & 0.00112 & ~~51.5~~ & 76.6508&
\textbf{0.00054} & ~~21.2~~ & 76.6468&
0.00056 & ~~22.2~~ & 76.6585\\
\hline $100\times 100$ & 0.00538 & ~~51.1~~ & 217.001& 0.00252 &
~~22.2~~ & 216.985&
\textbf{0.00225} & ~~22.9~~ & 216.987\\
\hline $300\times 300$ & 0.1859 & ~~50~~ & 1123.66& 0.10501 &
~~21.9~~ & 1123.67&
\textbf{0.08792} & ~~22.9~~ & 1123.70\\
\hline $500\times 500$ & 1.3802 & ~~50~~ & 2420.41& 0.6574 &
~~22~~ & 2419.63&
\textbf{0.5343} & ~~21~~ & 2420.28\\
\hline $1000\times 1000$ & 53.0936 & ~~50~~ & 6844.82& 25.8139 &
~~23~~ & 6844.14&
\textbf{18.5953} & ~~23~~ & 6845.32\\
\hline
\end{tabular}
}
\vspace{-0.5cm}
\end{center}
\end{table*}

\section{Experiments}\label{sec:exper}
In this section, we conduct several experiments on both the inner
loop only and the complete TILT problem to evaluate the
performance of our proposed LADMAP with warm starts method.
Numerical experiments on the inner loop only are conducted by
using synthetic data in order to demonstrate the effectiveness of
LADMAP and the warm start for SVD. For the complete TILT problem,
we conduct experiments on images with simulated and real
deformations to further test the efficiency and robustness of
LADMAP and the warm start techniques. The images we use are from a
low-rank textures data set and a natural images data set,
respectively.

The code for the original ADM based method is provided by the
first author of~\citep{ZhangTILT:11}. For fair comparison, we set
the common parameters in all compared methods the same. All the
codes are in MATLAB and the experiments are run on a workstation
with an Intel Xeon E5540@2.53GHz CPU and 48GB memory.

\subsection{Numerical Study on the Inner Loop Only}
In this section, we use synthetic data to compare the efficiency
of the original ADM based algorithm, our LADMAP based algorithm,
and LADMAP with warm start for computing SVD (LADMAP+SVDWS) on the
inner loop only. As this time we only focus on the inner loop, the
effectiveness of variable warm start, which requires outer loops,
cannot be shown. We generate $D\circ\tau$ and Jacobian $J$
randomly and use them as the input for ADM, LADMAP, and
LADMAP+SVDWS. For fair comparison, we tune the extra parameters
$\varepsilon_2$ and $\rho_0$ in LADMAP and LADMAP+SVDWS and
$\varepsilon_{svd}$ in LADMAP+SVDWS so that the three methods stop
with almost the same objective function values. All methods are
initialized as $A^0_0 = D\circ\tau^0$ and $E^0_0 = 0$.
$\Delta\tau_0$ in the ADM method is initialized as $\Delta\tau_0 =
0$. Under these conditions, we compare the three methods on their
computation time, the number of iterations needed to converge, and
the objective function value when the iterations stop. The
comparison is done under different sizes of $D\circ\tau$. All the
tests are repeated for 10 times and the average quantities are
reported.

The comparative results are shown in Table~\ref{tab:numerical}. We
can observe that all the three methods arrive at roughly the same
objective function values. However, LADMAP uses less than half of
the number of iterations than ADM does and the SVD warm start only
changes the number of iterations very slightly. Consequently,
LADMAP is much faster than ADM, while SVDWS further speeds up the
computation of LADMAP when the size of $D\circ\tau$ is not too
small (e.g., $\ge 100$). The acceleration rate also increases when
the size of $D\circ\tau$ grows\footnote{We will see much higher
acceleration rates when using SVDWS on real data (see
Section~\ref{sec:Compare_Speed}).}. When the size of $D\circ\tau$
is very small (e.g., $<100$), SVDWS does not seems to speed up the
computation. This may due to the ultra-short computing time and
hence other processes on the workstation supporting the computing
environment can influence the total computing time. Anyway, the
slowdown is rather insignificant. As the speed of solving large
sized TILT is more time demanding in real applications, adopting
SVDWS is still advantageous.

\subsection{Comparisons on the Complete TILT Problem}
In this subsection, using real image data we compare the original
ADM method, LADMAP, LADMAP with variable warm start (LADMAP+VWS),
and LADMAP  with both variable warm start and SVD warm start
(LADMAP+VWS+SVDWS) on solving the whole TILT problem, in order to
show the effectiveness of LADMAP and the two warm start
techniques. As we have pointed out before that the original ADM
may not converge to the optimal solution of the inner loop, we
first compare the convergence performance of ADM and LADMAP. Then
we test the robustness of ADM and LADMAP when there are
corruptions. We also present some examples on which ADM fails but
our LADMAP works well. Finally, we compare the computation time of
various methods on both synthetically and naturally deformed
images. The synthetically deformed images are generated by
deforming the low-rank textures with predefined transforms. The
naturally deformed images are from our images data set which
contains over 100 images downloaded from the web.

\begin{figure}[t]
\begin{center}
\includegraphics[width=8cm]{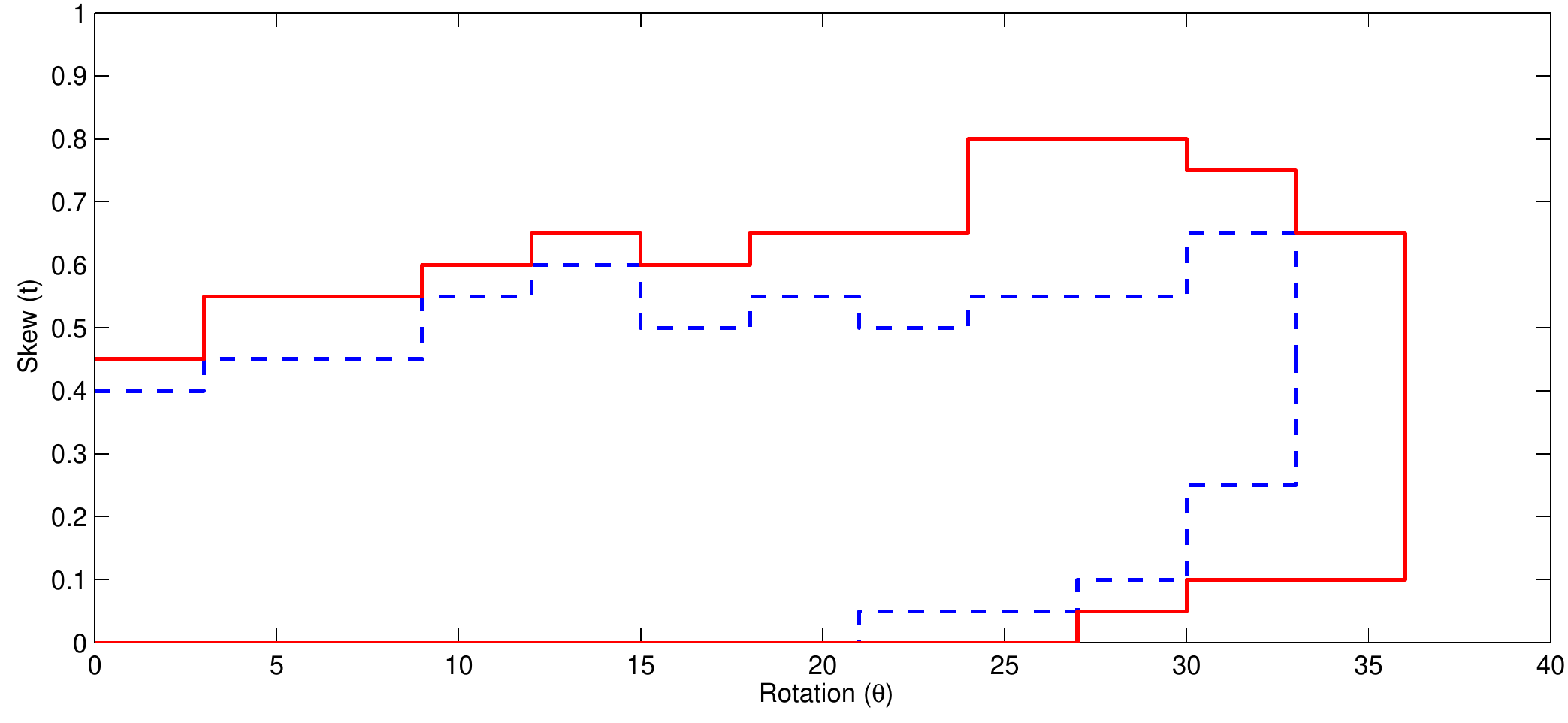}
\caption{\textbf{Range of convergence for affine transform.} The
x-axis and y-axis stand for the rotation angle $\theta$ and the
skew value $t$ in an affine transform, respectively. Regions
inside the curves are affine transforms that are successfully
recovered in all trials. The blue dashed curve and the red solid
curve are the boundaries of the parameters of affine transforms
that can be successfully recovered by ADM and LADMAP,
respectively.} \label{fig:convergence}
\end{center}
\vspace{-0.7cm}
\end{figure}

\begin{figure}[h]
\centerline{ \subfigure{
\includegraphics[width=0.25\columnwidth]{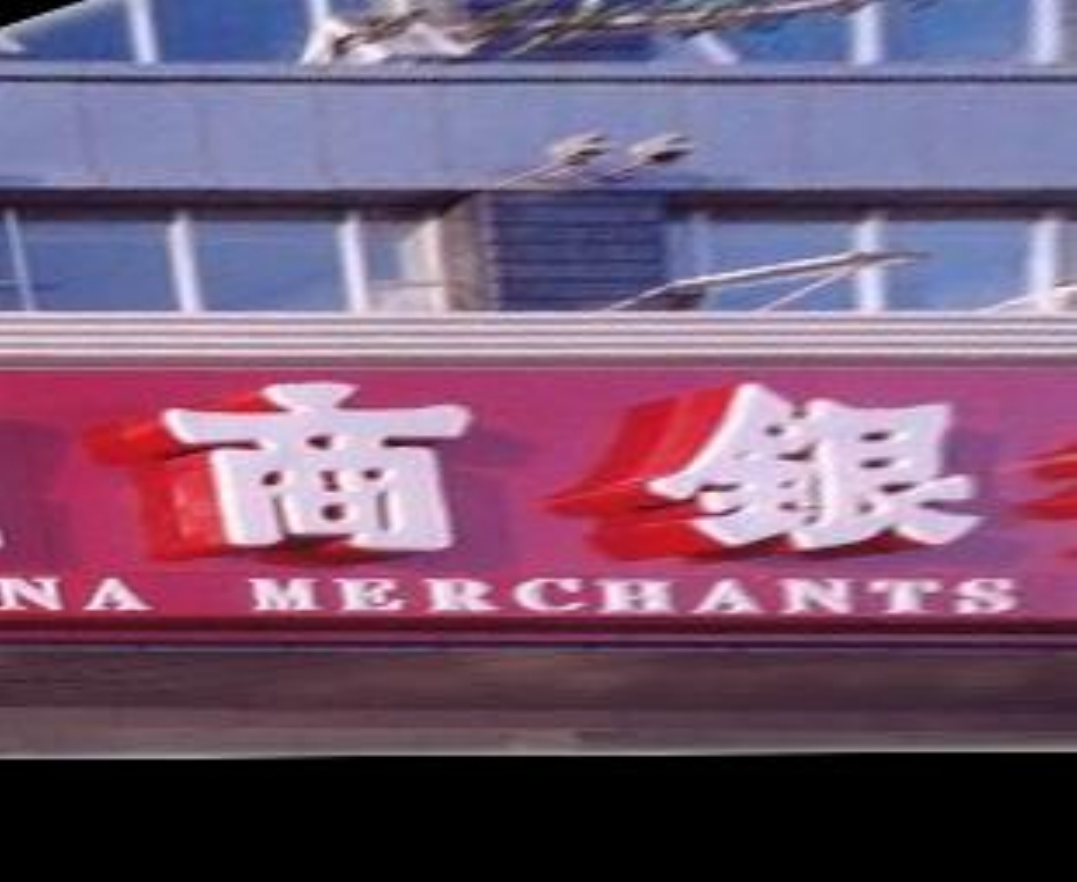}
} \subfigure{
\includegraphics[width=0.25\columnwidth]{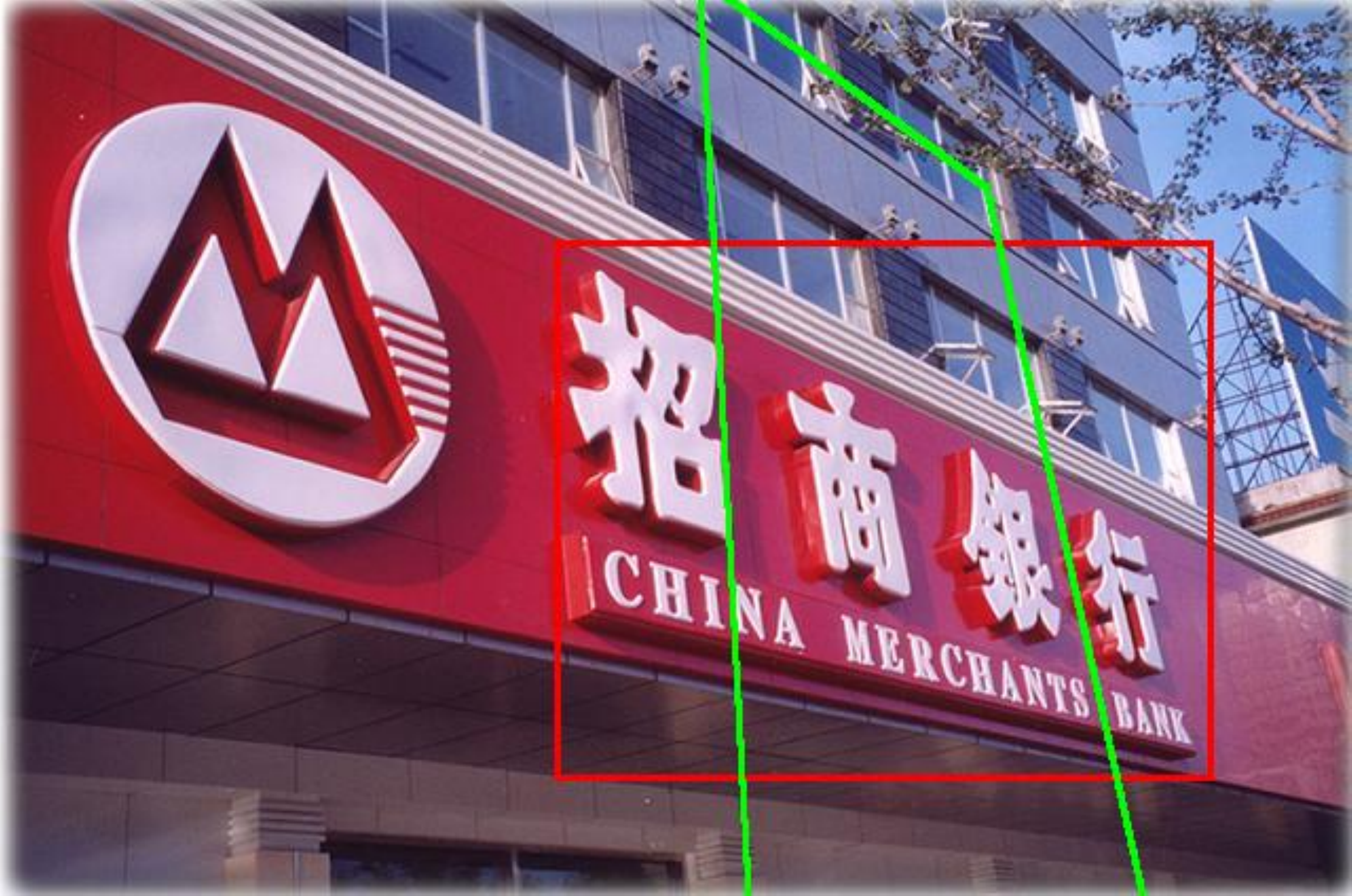}
} \subfigure{
\includegraphics[width=0.25\columnwidth]{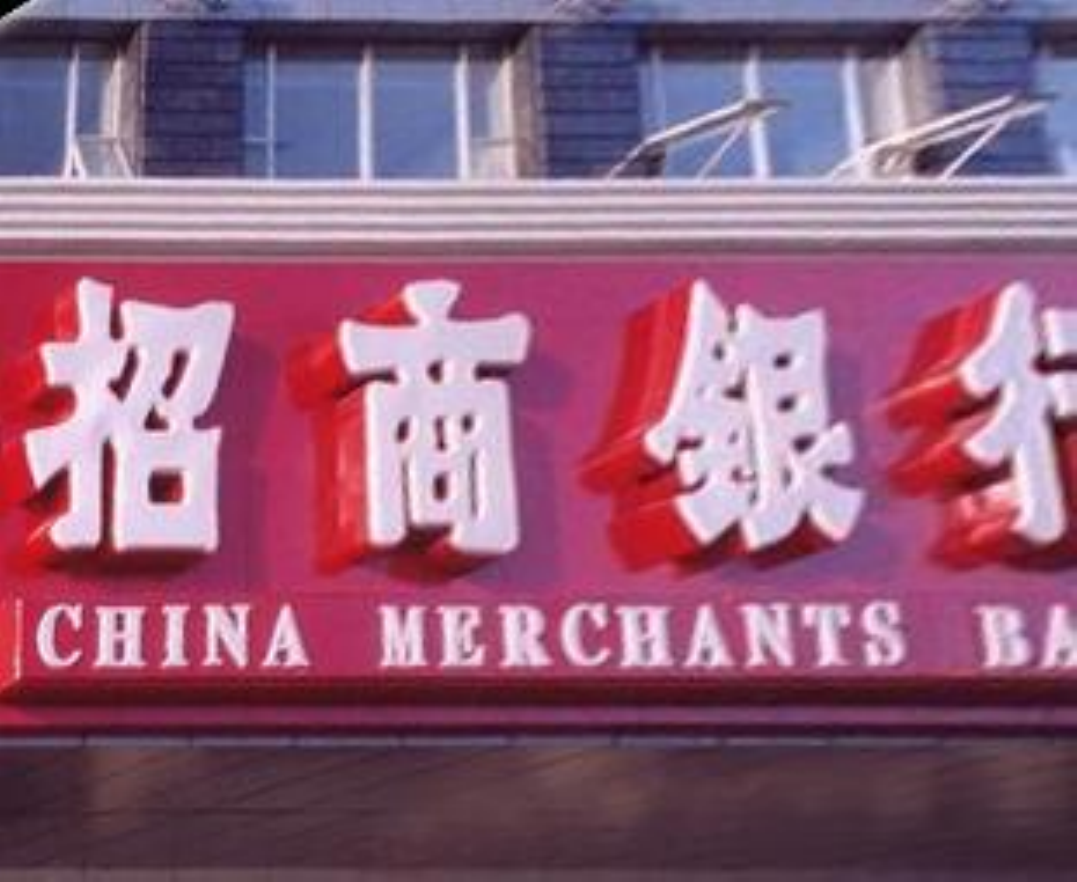}
} \subfigure{
\includegraphics[width=0.25\columnwidth]{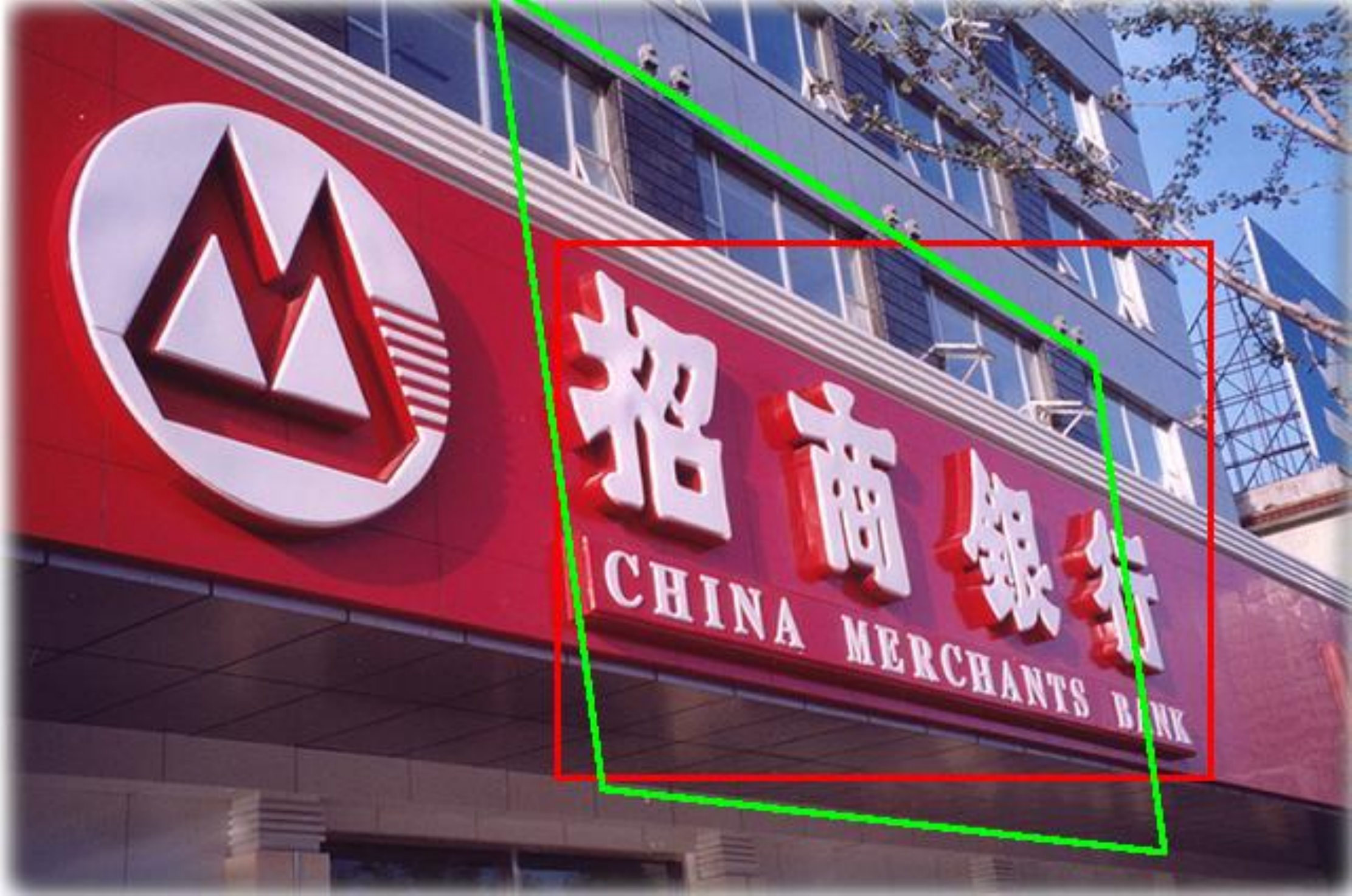}
} }
\centerline{ \subfigure{
\includegraphics[width=0.25\columnwidth]{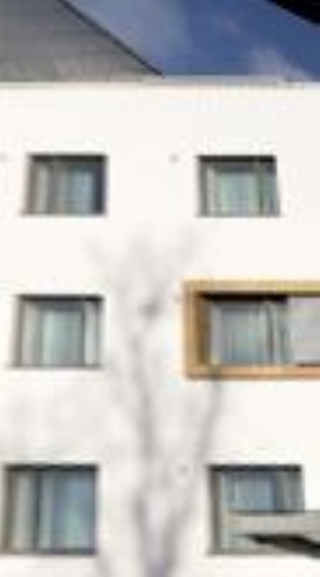}
} \subfigure{
\includegraphics[width=0.25\columnwidth]{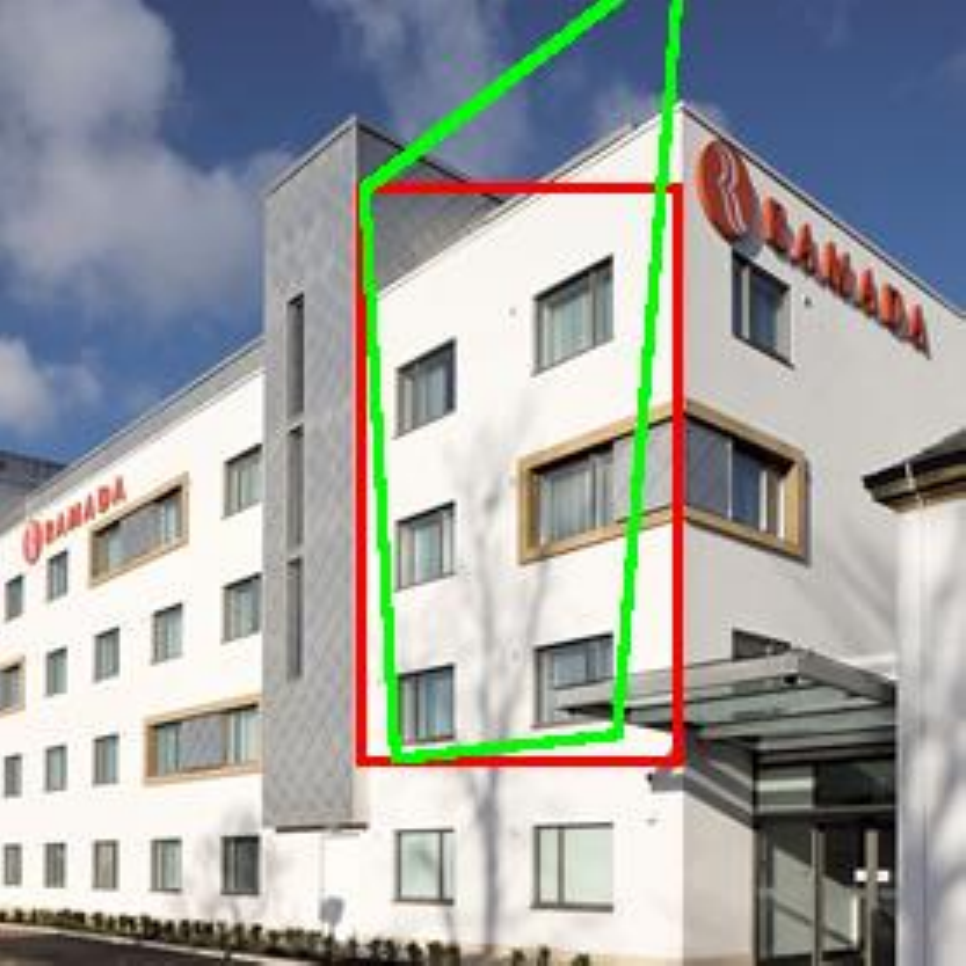}
} \subfigure{
\includegraphics[width=0.25\columnwidth]{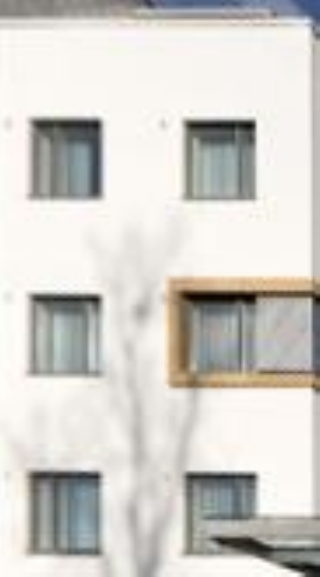}
} \subfigure{
\includegraphics[width=0.25\columnwidth]{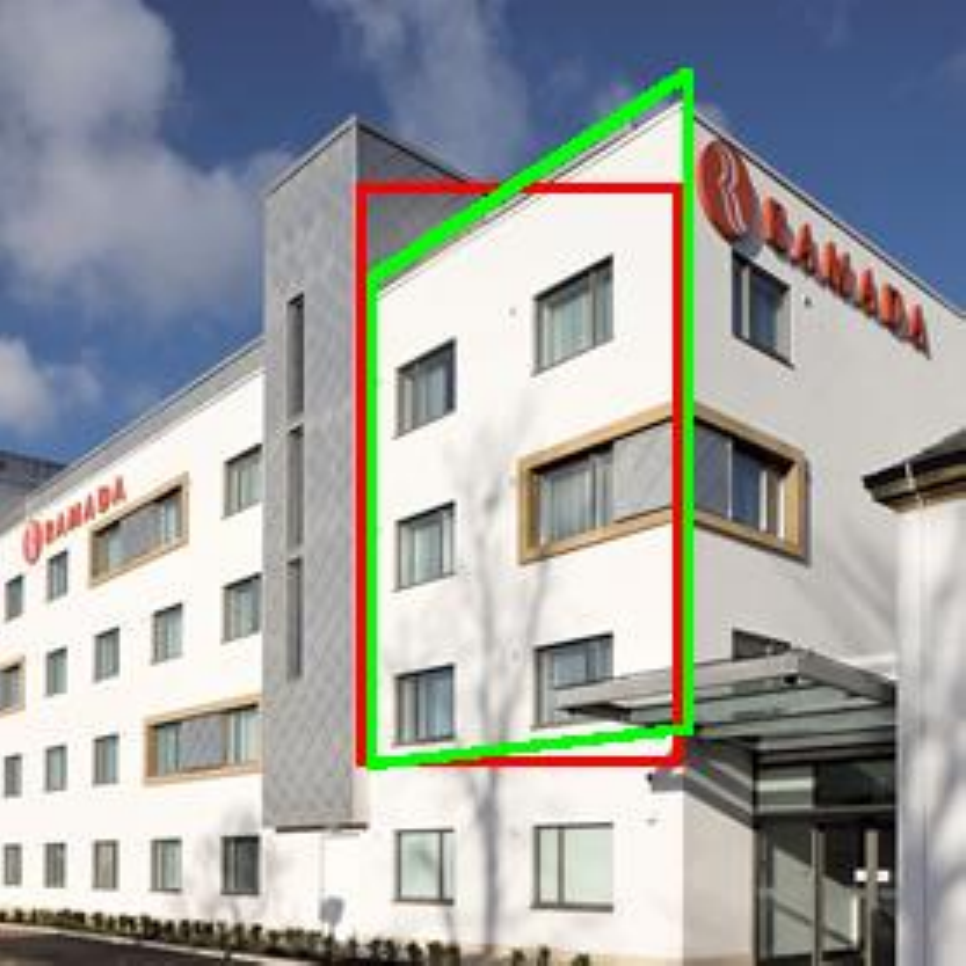}
} } \centerline{ \subfigure{
\includegraphics[width=0.25\columnwidth]{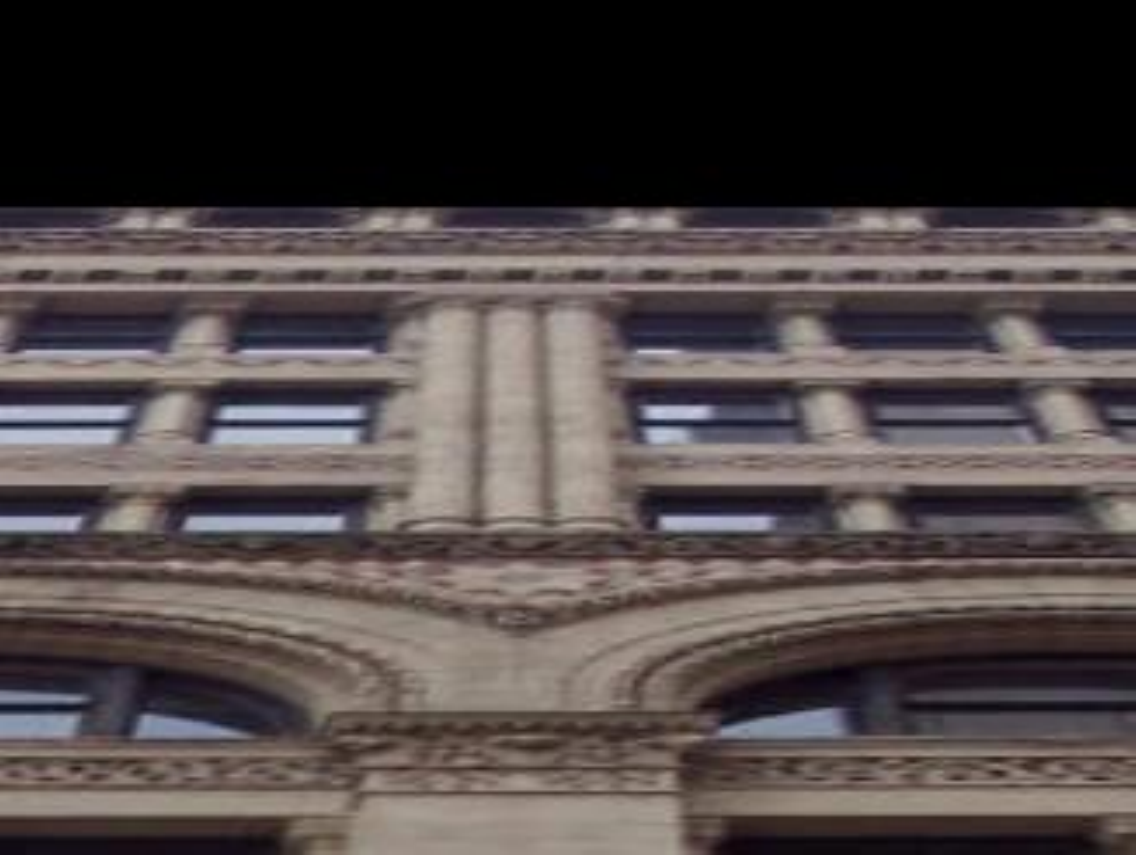}
} \subfigure{
\includegraphics[width=0.25\columnwidth]{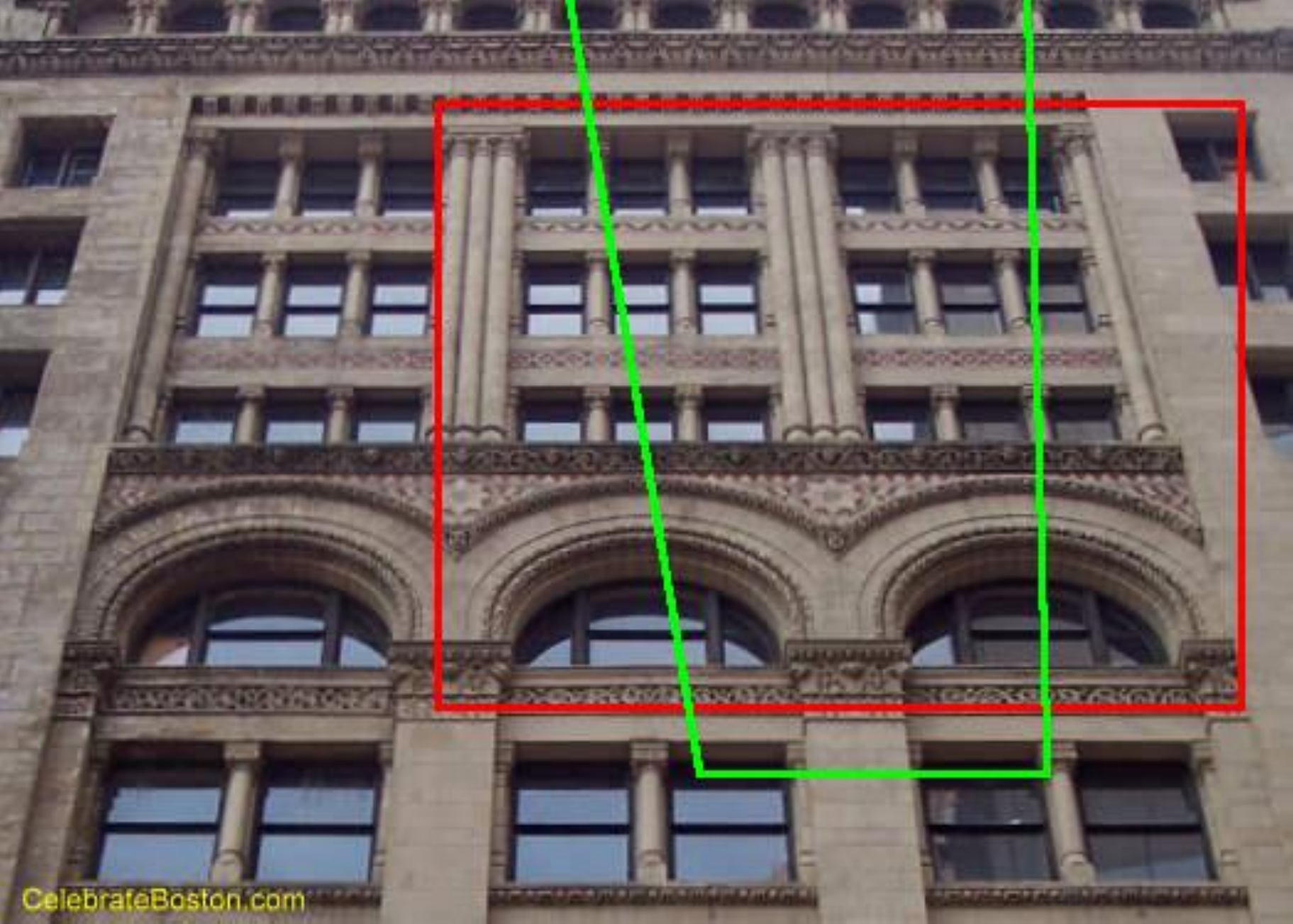}
} \subfigure{
\includegraphics[width=0.25\columnwidth]{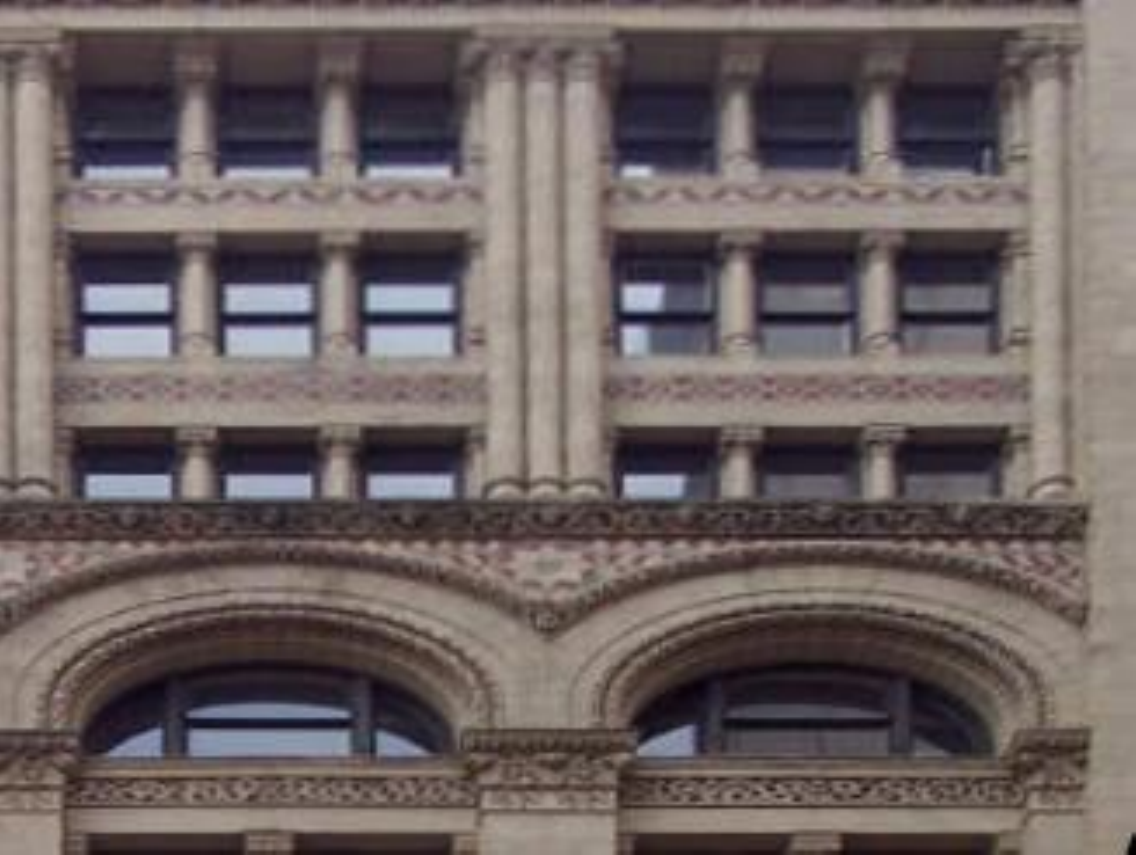}
} \subfigure{
\includegraphics[width=0.25\columnwidth]{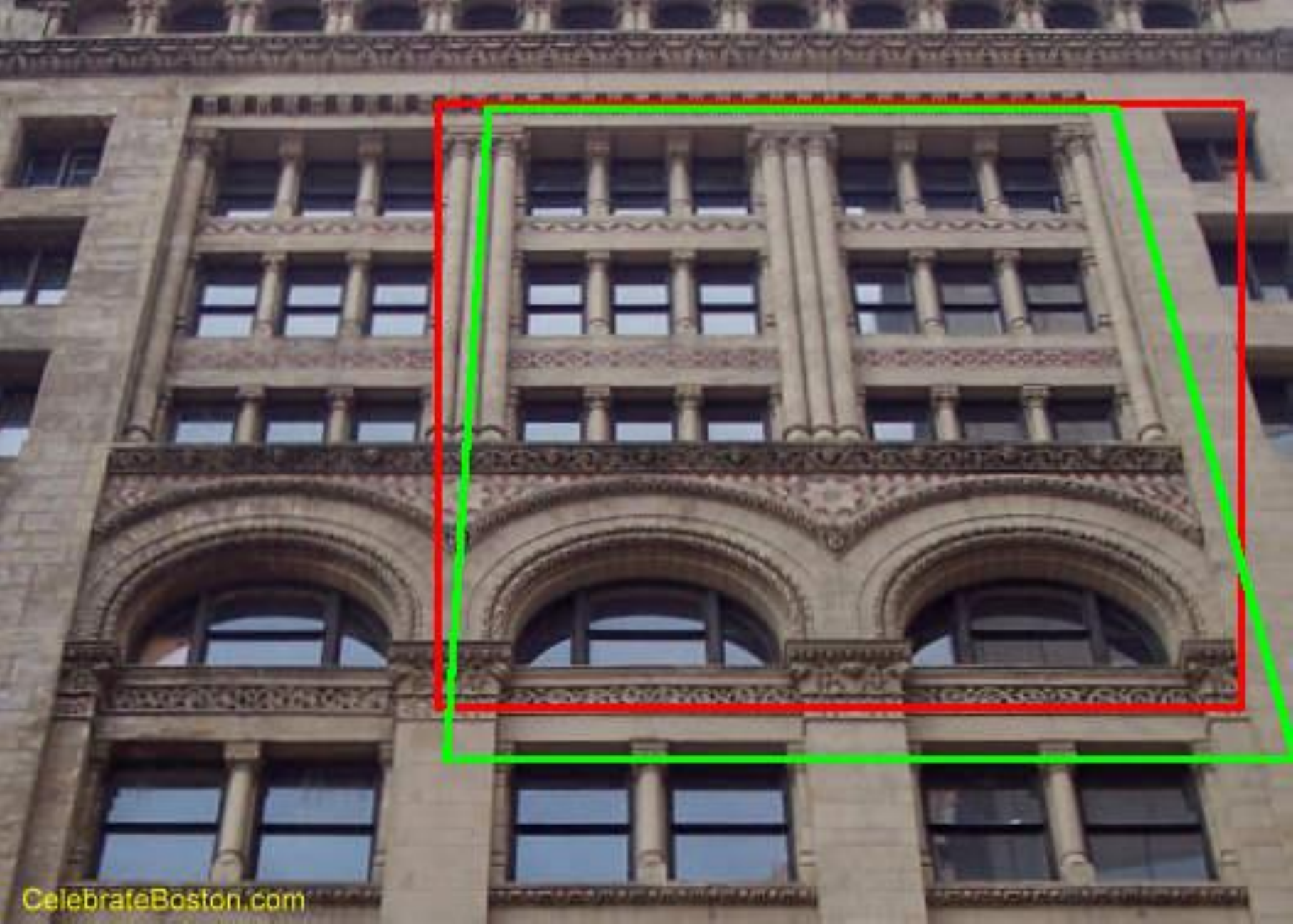}
} }
\centerline{ \subfigure{
\includegraphics[width=0.25\columnwidth]{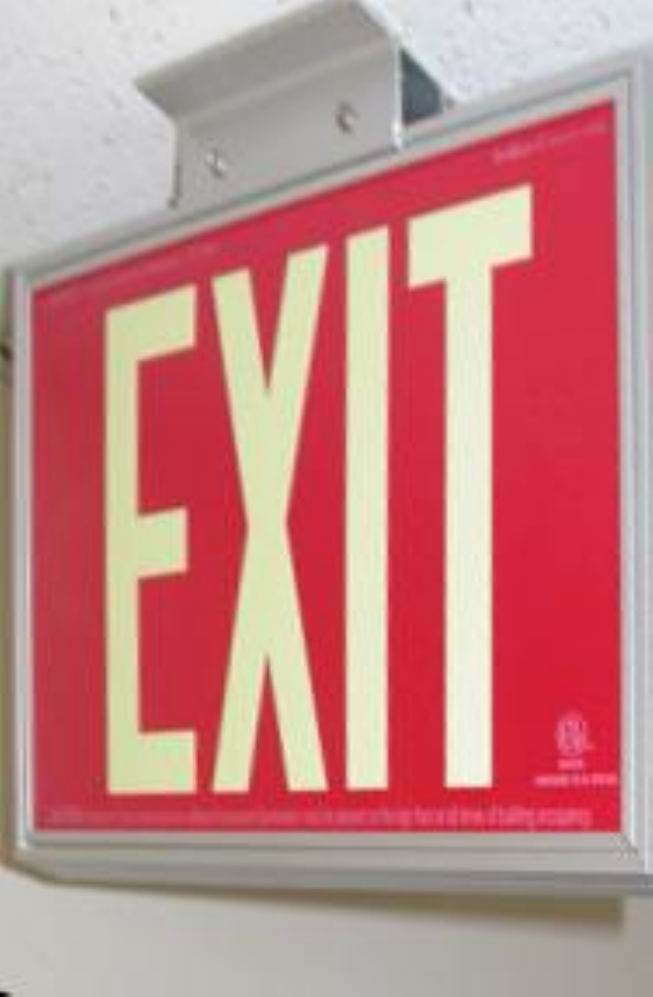}
} \subfigure{
\includegraphics[width=0.25\columnwidth]{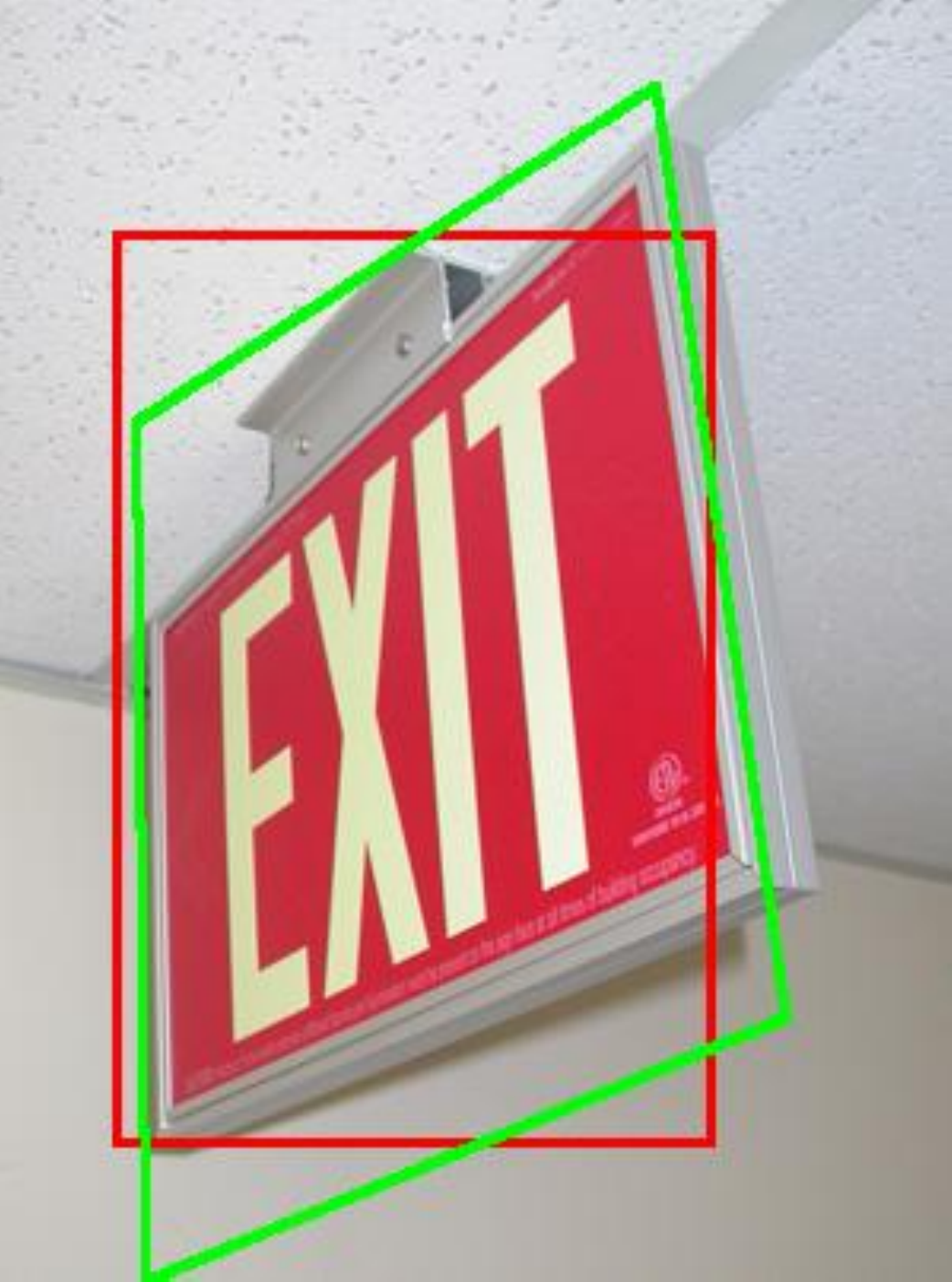}
} \subfigure{
\includegraphics[width=0.25\columnwidth]{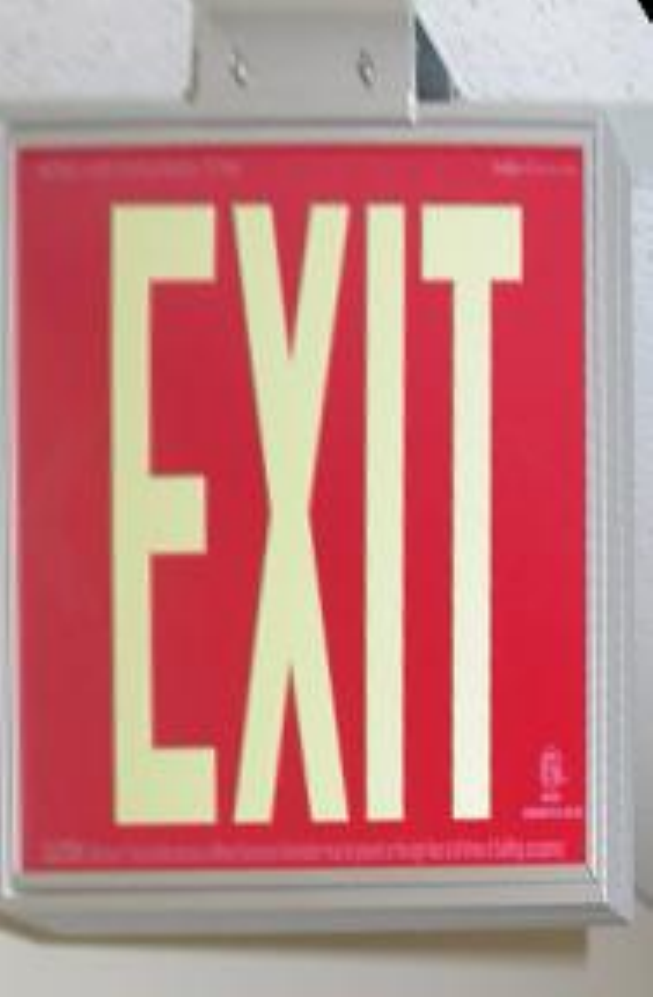}
} \subfigure{
\includegraphics[width=0.25\columnwidth]{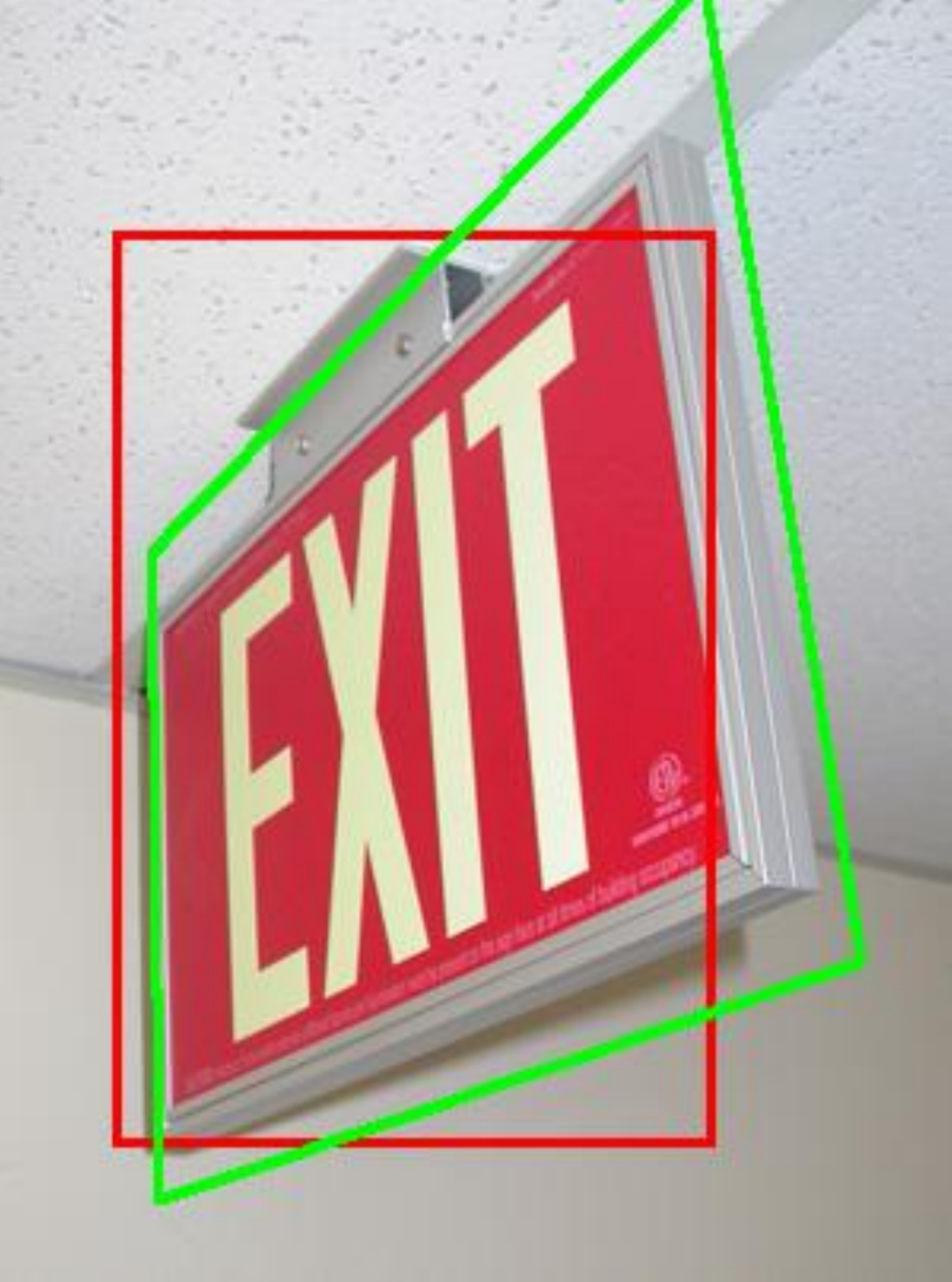}
} } \caption{{\bf Visual comparison between the results by ADM and
LADMAP.} First and second columns: rectification results by ADM.
Third and fourth columns: results by LADMAP. First and third
columns: rectified regions of interest. Second and fourth columns:
the initial transform (illustrated by red rectangles which are
manually prescribed) and the computed transform (illustrated by
green quadrilaterals).} \label{fig:empirical}
\end{figure}

\paragraph{Range of Convergence}
Since LADMAP for the inner loop is proven to converge to the
optimal solution, we expect that it will outperform ADM in
recovering the correct deformation when solving the whole TILT
problem\footnote{Note that as the whole TILT problem
(\ref{Eq:TILT_2}) is \emph{not} a convex program, LADMAP cannot
achieve the global optimum either. So LADMAP can also fail to
recover the correct deformation.}. To show that LADMAP can recover
broader range of deformation than ADM does, we test them with a
standard checker-board pattern.

Following the same setting in~\citep{ZhangTILT:11}, we deform a
checker-board pattern by an affine transform: $y=Ax+b$, where
$x,y\in \mathbb{R}^2$, and test if the two algorithms can recover
the correct transform. The matrix $A$ is parameterized as
$A(\theta,t)=$ $\left[
                 \begin{array}{cc}
                   \mathrm{cos}~\theta & -\mathrm{sin}~\theta \\
                   \mathrm{sin}~\theta & \mathrm{cos}~\theta \\
                 \end{array}
               \right]
        \left[
          \begin{array}{cc}
            1 & t \\
            0 & 1 \\
          \end{array}
        \right]$, where $\theta$ is the rotation angle and $t$ is the skew value.
We change $\theta$ within the range $[0, \pi/6]$ with a step size
$\pi/60$, and $t\in [0,1]$ with a step size 0.05. We can observe
from Fig.~\ref{fig:convergence} that the range of convergence of
our LADMAP completely encloses that of ADM. So the working range
of LADMAP is larger than that of ADM.

We further test with real images taken from natural scenes and
manually prescribe the regions to be rectified. We have found many
examples that LADMAP works better than ADM. However, we have not
encountered any example that ADM works better. Part of the
examples are shown in Fig.~\ref{fig:empirical}. They are chosen
according to the challenging cases listed in \citep{ZhangTILT:11}
for TILT to rectify. For example, the first example is lacking
regularity in the printed texts or the prescribing rectangular
contains too much background; the second example has very large
deformation; and the third example may has a boundary effect. On
these examples, LADMAP works much better than ADM in rectifying
the prescribed regions.

\begin{figure*}
\centering{
\hspace{-0.2cm}
    \subfigure[]{
        \includegraphics[width=0.95\columnwidth]{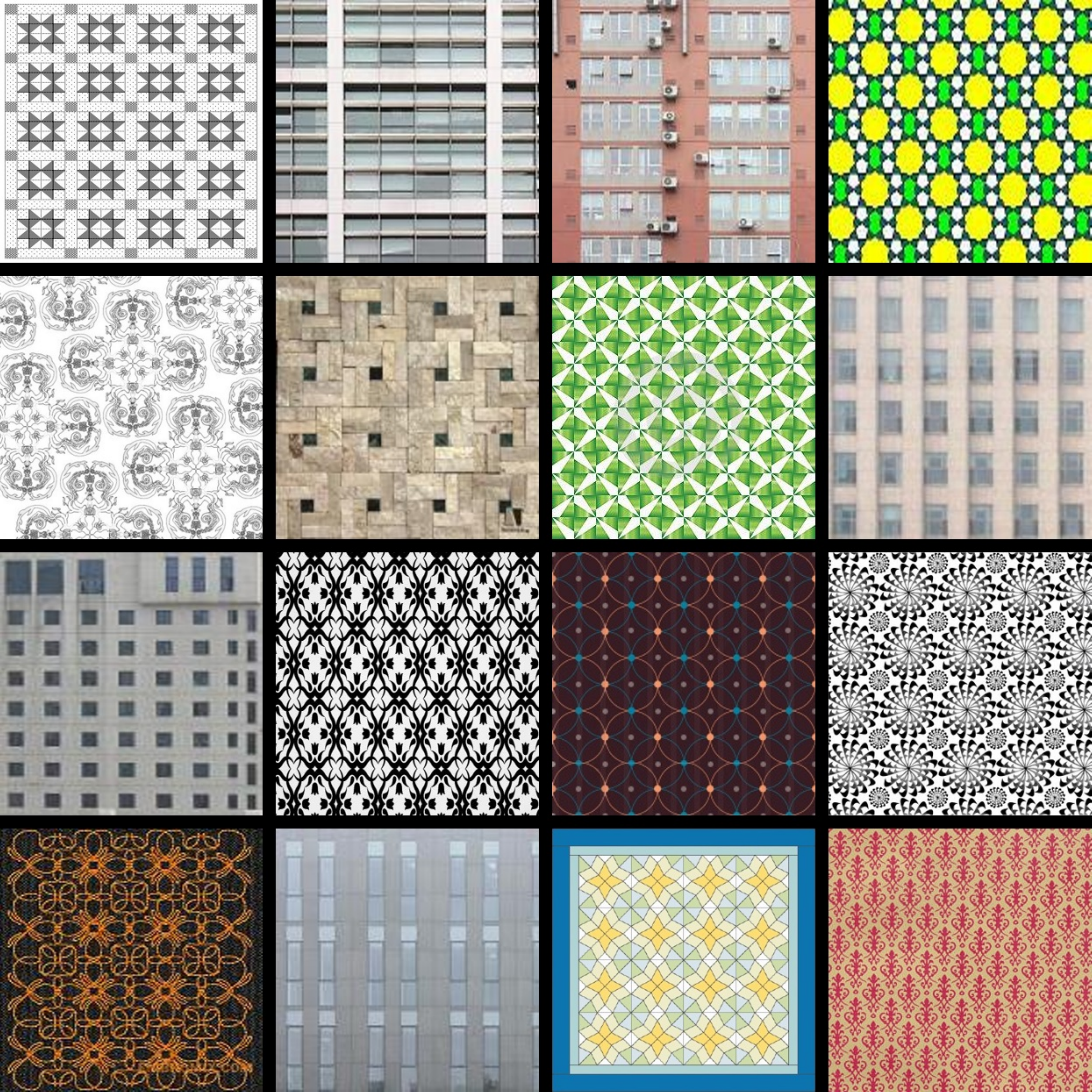}}
\hspace{-0.2cm}
    \subfigure[]{
        \includegraphics[width=0.95\columnwidth]{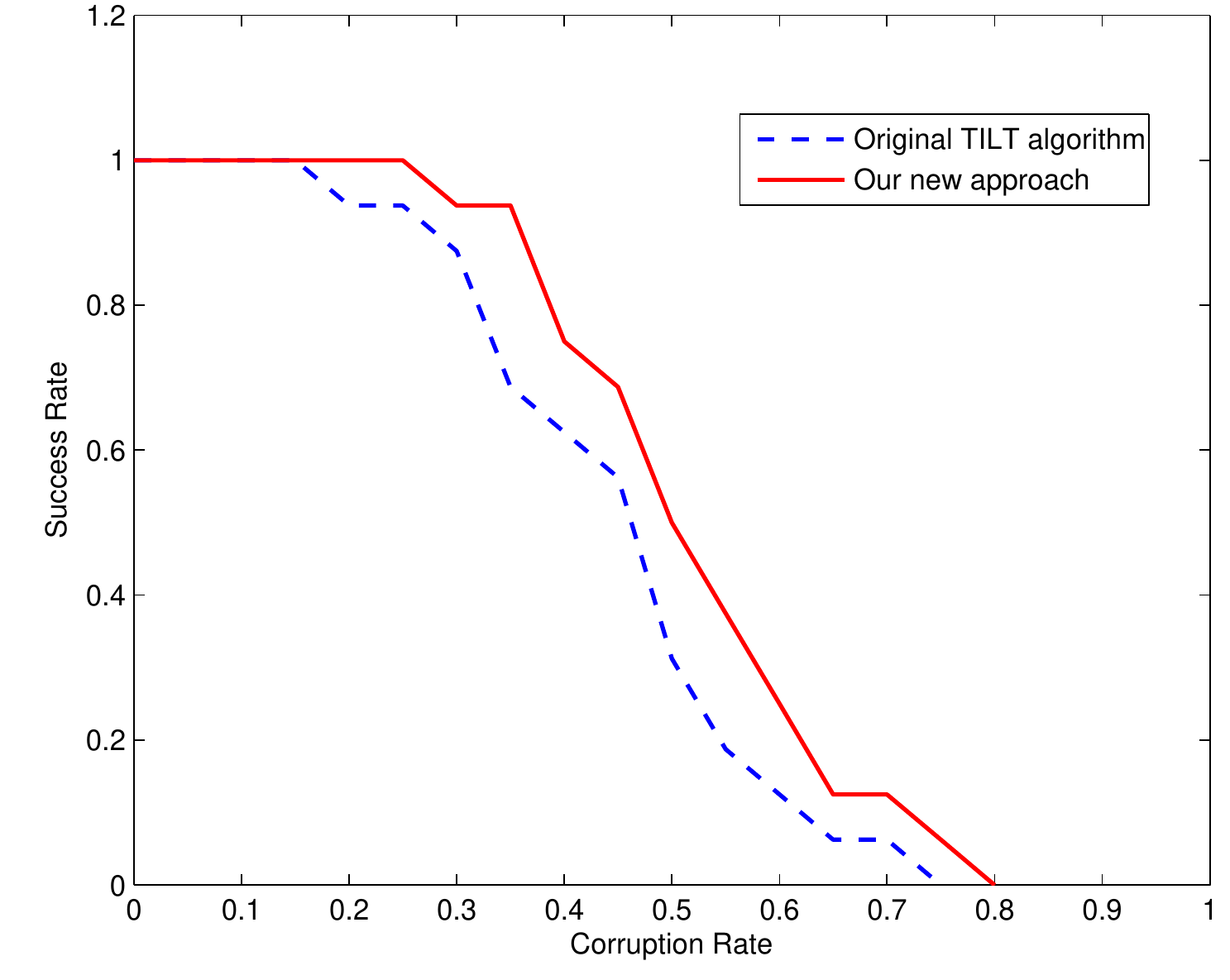}}}
\caption{\textbf{Robustness of ADM and LADMAP under different
levels of corruption.} (a) Sample textures. (b) Success rates of
ADM and LADMAP under different levels of corruption.}
\label{fig:robustness}
\end{figure*}

\paragraph{Robustness under Corruption} 
In this subsection, we compare the robustness of ADM and LADMAP
when there is corruption in images. We test with some low-rank
textures shown in Fig. \ref{fig:robustness}(a). Following the same
setting in \citep{ZhangTILT:11}, we introduce a small deformation,
say rotation by $10^\circ$, and examine whether the two methods
can recover the correct transform under different levels of random
corruption. We randomly select a fraction (from 0\% to 100\%) of
the pixels and assign all their RGB values as random values
uniformly distributed in $(0, 255)$. We run the two methods on
such corrupted images and record at each level of corruption how
many images are correctly rectified.

The comparative results are shown in Fig. \ref{fig:robustness}(b).
We can see that when the percentage of corruption is larger than
15\%, our LADMAP always outperforms ADM. For example, when 50\% of
the pixels are corrupted, LADMAP can still obtain the correct
solution on more than half of the test images, while ADM can only
deal with about 30\% of the images. The above comparison
demonstrates that our LADMAP method is more robust than the
original ADM method when there is corruption in images.

\begin{table*}[t]
\begin{center}
\caption{\textbf{Comparison of speed on images with artificial
affine deformations.} The time costs are counted in seconds. The
relative errors are between the computed $\tau$ and the ground
truth.} \label{tab:speed_affine} \vspace{0.1cm} {\small
\begin{tabular}{|c|c||c|c|c|c|c|c|c|c|}
\hline
 &
 & \multicolumn{2}{|c|}{ADM}
 & \multicolumn{2}{|c|}{LADMAP}
 & \multicolumn{2}{|c|}{LADMAP+VWS}
 & \multicolumn{2}{|c|}{LADMAP+VWS+SVDWS}\\
\hline \hline \textbf{$\theta=30^\circ$}
 & size
&time (s) & rel. err. & time (s) & rel. err. & time (s) & rel. err. & time (s) & rel. err. \\
\hline 1 & 50$\times$43& 6.6318 & 0.1563 & 2.2941 & 0.1561 &
1.1857 & 0.1561 &
0.4034 & 0.1560 \\
\hline
2 & 31$\times$36&
0.2503 & 0.0086 &
0.1834 & 0.0087 &
0.0913 & 0.0085 &
0.0063 & 0.0085 \\
\hline
3 & 38$\times$37&
0.1594 & 0.00087 &
0.1057 & 0.00087 &
0.0459 & 0.00088 &
0.0338 & 0.00088 \\
\hline
4 & 28$\times$26&
0.1091 & 0.0040 &
0.0634 & 0.0040 &
0.0408 & 0.0039 &
0.0353 & 0.0039 \\
\hline
5 & 38$\times$45&
3.139 & 0.4226 &
1.977 & 0.4219 &
0.420 & 0.4193 &
0.199 & 0.4191 \\
\hline\hline \textbf{$t=0.2$}
 & size &
time (s) & rel. err. & time (s) & rel. err. & time (s) & rel. err. & time (s) & rel. err. \\
\hline
1 & 27$\times$27&
0.8926 & 0.0252 &
0.3068 & 0.0245 &
0.1605 & 0.0243 &
0.0995 & 0.0247 \\
\hline
2 & 47$\times$50&
3.2381 & 0.2066 &
2.1583 & 0.2082 &
0.4551 & 0.1944 &
0.1988 & 0.1949 \\
\hline
3 & 47$\times$45&
0.6271 & 0.0065 &
0.5043 & 0.0065 &
0.3914 & 0.0067 &
0.3124 & 0.0066 \\
\hline
4 & 30$\times$26&
2.4916 & 0.1896 &
1.3644 & 0.1881 &
0.9039 & 0.1859 &
0.6126 & 0.1849 \\
\hline
5 & 26$\times$28&
0.3681 & 0.0538 &
0.2254 & 0.0536 &
0.1622 & 0.0536 &
0.1274 & 0.0539 \\
\hline
\end{tabular}
}

\vspace{1em}

\caption{\textbf{Comparison of computation speed on real images.}
The time costs are counted in second. The speedup rates are w.r.t.
the baseline ADM method.} \label{tab:speed_natural} \vspace{0.1cm}
{\small
\begin{tabular}{|c|c||c|c|c|c|c|c|c|}
\hline
\textbf{Case}
 & size
 & \multicolumn{1}{|c|}{~~~ADM~~~}
 & \multicolumn{2}{|c|}{LADMAP}
 & \multicolumn{2}{|c|}{LADMAP+VWS}
 & \multicolumn{2}{|c|}{LADMAP+VWS+SVDWS}\\
\hline \hline &  & time (s) & time (s) & speedup & time (s) &
speedup &
time (s) & speedup \\
\hline 1 & 82$\times$82& 6.2728 & 4.8299 & 1.29 & 1.8119 & 3.46 &
1.1386 & 5.50 \\
\hline
2 & 84$\times$78&
7.6238 &
2.8338 & 2.69 &
1.6067 & 4.75 &
1.3056 & 5.84 \\
\hline
3 & 80$\times$87&
5.4040 &
2.4863 & 2.17 &
1.3238 & 4.08 &
0.9587 & 5.64 \\
\hline
4 & 68$\times$72&
4.7575 &
2.8580 & 1.66 &
0.9953 & 4.78 &
0.8237 & 5.78 \\
\hline
5 & 51$\times$60&
3.1128 &
2.7048 & 1.15 &
0.8821 & 3.53 &
0.4906 & 6.35 \\
\hline
6 & 89$\times$90&
3.3004 &
2.2573 & 1.46 &
0.8274 & 3.99 &
0.5174 & 6.38 \\
\hline
7 & 96$\times$107&
6.7319 &
2.2305 & 3.02 &
1.6650 & 4.04 &
0.8212 & 8.20 \\
\hline
8 & 65$\times$60&
2.5367 &
1.4115 & 1.79 &
0.5022 & 5.05 &
0.4172 & 6.08 \\
\hline
9 & 72$\times$71&
2.5501 &
0.9307 & 2.73 &
0.6234 & 4.09 &
0.3967 & 6.43 \\
\hline
10 & 83$\times$90&
4.3284 &
3.6923 & 1.17 &
1.1078 & 3.91 &
0.7141 & 6.06 \\
\hline
\end{tabular}
}
\end{center}
\end{table*}

\paragraph{Speed of the Algorithms}\label{sec:Compare_Speed}
In this subsection, we report the computational cost of four
different algorithms: ADM, LADMAP, LADMAP+VWS, and LADMAP+VWS+
SVDWS for solving the whole TILT problem, in order to show the
computational improvement from each component.

The comparisons are done on two types of images. The first type of
images are obtained by applying affine transforms to real images.
The second one are images of natural scenes that can undergo
either affine or projective transform, in which the rectangular
regions to be rectified are manually prescribed.

For the first type of images, part of the comparison results,
namely the images are rotated with $\theta = 30^\circ$ only or
skewed with $t=0.2$ only, are presented in
Table~\ref{tab:speed_affine}. We tuned the four algorithms so that
all of them produced nearly the same relative error, which is
defined as $\frac{\|\tau^* - \tau_G\|}{\|\tau_G\|}$ where $\tau^*$
is the optimal solution produced by the corresponding algorithm
and $\tau_G$ is the ground-truth deformation matrix. We can see
that LADMAP is much faster than ADM, the variable warm start
speeds up the convergence, and SVD warm start further cuts the
computation cost.

For the second type of images, comparison of the speed on ten of
the images is shown in Table~\ref{tab:speed_natural}. We can see
that LADMAP can be at least 20\% faster than ADM, with the
variable warm start LADMAP can be at least 2.5 times faster than
ADM, and with the SVD warm start further added, LADMAP can be at
least 4.5 times faster than ADM. So the speed advantage of our new
algorithm is apparent.

\section{Conclusions}\label{sec:concl}
In this paper we propose an efficient and robust algorithm for
solving the recently popular TILT problem. We reformulate the
inner loop of TILT and apply a recently proposed LADMAP to solve
the subproblem of the inner loop. For further speed up, we also
propose variable warm start for initialization and introduce an
SVD warm start technique to cut the computational cost of
computing SVDs. Extensive experiments have testified to the better
convergence property, higher robustness to corruption, and faster
speed of LADMAP, and the effectiveness of our warm start
techniques.

\begin{acknowledgements}
Z. Lin is supported by the National Natural Science Foundation of
China (Grant No.s 61272341 and 61231002).
\end{acknowledgements}

\section*{Appendix A: Details of Computing SVD with Warm Start}
In this section, for brevity we simply write $U(0)$, $\Sigma(0)$,
and $V(0)$ as $U$, $\Sigma$, and $V$, respectively.
\subsection{Gradients}
The gradients are:
\begin{eqnarray}
  G_U &=& \frac{\partial F}{\partial U} = U\Sigma^2-MV\Sigma, \label{Eq:gradient1}\\
  G_V &=& \frac{\partial F}{\partial V} = V\Sigma^2-M^TU\Sigma,\label{Eq:gradient2}\\
  G_\Sigma &=& \frac{\partial F}{\partial \Sigma} = \Sigma-U^TMV\label{Eq:gradient3}.
\end{eqnarray}
The projected gradients are:
\begin{small}
\begin{eqnarray}\label{Eq:projected_gradient}
  P_U &=& G_UU^T-UG_U^T=(USV^T)M^T-M(USV^T)^T, \label{Eq:projected_gradient1}\\
  P_V &=& G_VV^T-VG_V^T=(USV^T)^TM-M^T(USV^T),\label{Eq:projected_gradient2}\\
  P_\Sigma &=& \diag(\Sigma - U^TMV)\label{Eq:projected_gradient3}.
\end{eqnarray}
\end{small}
Note that $USV^T$ is just the SVD of the matrix in last iteration.
So we do not need to recompute their products.

\subsection{Derivatives of $f(t)$}
\subsubsection{First-order Derivative}
We utilize the chain rule to calculate the derivative, let
\begin{equation*}
    H = h(t) = M - U(t)\cdot \Sigma(t)\cdot V(t)^T
\end{equation*}
and $\ell(H)=\frac{1}{2}\|H\|_F^2$, then
\begin{eqnarray*}
  f'(t) = \frac{\mbox{d}  \ell(H)}{\mbox{d} t} &=& \tr \Big[ \left(\frac{\mbox{d} \ell(H)}{\mbox{d} H}\right)^T \cdot \frac{\mbox{d}  h(t)}{\mbox{d} t} \Big] \\
    &=& \tr \Big[ h(t)^T\cdot h'(t) \Big],
\end{eqnarray*}
where
\begin{equation*}
\begin{array}{l}
    h'(t) = -U'(t)\Sigma(t)V(t)^T - U(t)\Sigma'(t)V(t)^T \\
    \quad\quad- U(t)\Sigma(t)V'(t)^T,
\end{array}
\end{equation*}
and
\begin{equation*}
\begin{array}{l}
  \vspace{+1mm}
  U'(t) = -(I+\frac{1}{2}tP_U)^{-1}\frac{1}{2}P_U(I+\frac{1}{2}tP_U)^{-1}(I-\frac{1}{2}tP_U)U\\
  \qquad\qquad - (I+\frac{1}{2}tP_U)^{-1}\frac{1}{2}P_UU, \vspace{+3mm}\\
  \Sigma'(t) = -P_\Sigma, \vspace{+3mm}\\
  \vspace{+1mm}
  V'(t)^T = V^T\frac{1}{2}P_V(I-\frac{1}{2}tP_V)^{-1}\\
  \qquad\qquad +V^T(I+\frac{1}{2}tP_V)(I-\frac{1}{2}tP_V)^{-1}\frac{1}{2}P_V(I-\frac{1}{2}tP_V)^{-1}.
\end{array}
\end{equation*}

\subsubsection{Second-order Derivative}
Since $\mbox{d} \big(\tr(Q)\big) =\tr(\mbox{d}  Q)$ and $\mbox{d}
Q^T = (\mbox{d} Q)^T$, we have
\begin{equation*}\label{Eq:second}
    f''(t) = \frac{\mbox{d} f'(t)}{\mbox{d} t} = \tr\Big[ h'(t)^T\cdot h'(t) + h(t)^T\cdot
    h''(t)\Big],
\end{equation*}
where
\begin{equation*}
\begin{array}{l}
 h''(t) = \frac{\mbox{d} h'(t)}{\mbox{d} t}\\
  = -U''(t)\Sigma(t)V(t)^T - U'(t)\Sigma'(t)V(t)^T - U'(t)\Sigma(t)V'(t)^T\\
  - U'(t)\Sigma'(t)V(t)^T - U(t)\Sigma''(t)V(t)^T - U(t)\Sigma'(t)V'(t)^T\\
  - U'(t)\Sigma(t)V'(t)^T - U(t)\Sigma'(t)V'(t)^T - U(t)\Sigma(t)V''(t)^T
\end{array}
\end{equation*}
and
\begin{small}
\begin{equation*}
\begin{array}{l}
  U''(t) = \frac{\mbox{d} U'(t)}{\mbox{d} t} \vspace{+1mm}\\
  =\frac{1}{2}(I+\frac{1}{2}tP_U)^{-1}P_U(I+\frac{1}{2}tP_U)^{-1}P_U(I+\frac{1}{2}tP_U)^{-1}(I-\frac{1}{2}tP_U)U\vspace{+1mm}\\
  \quad + \frac{1}{2}(I+\frac{1}{2}tP_U)^{-1}P_U(I+\frac{1}{2}tP_U)^{-1}P_UU, \vspace{+3mm}\\
  \Sigma''(t) = 0,\vspace{+3mm}\\
  V''(t)^T = \frac{\mbox{d} V'(t)}{\mbox{d} t}\vspace{+1mm}\\
  =\frac{1}{2}V^T(I+\frac{1}{2}tP_V)(I-\frac{1}{2}tP_V)^{-1}P_V(I-\frac{1}{2}tP_V)^{-1}P_V(I-\frac{1}{2}tP_V)^{-1}\vspace{+1mm}\\
  \quad +\frac{1}{2}V^TP_V(I-\frac{1}{2}tP_V)^{-1}P_V(I-\frac{1}{2}tP_V)^{-1}.
\end{array}
\end{equation*}
\end{small}

\subsection{Taylor Expansion}
We want to do the second-order Taylor expansion of $f(t)$ at
$t=0$:
$$f(t) \approx f(0)+f'(0)\cdot t + \frac{1}{2}f''(0)\cdot t^2.$$
First, the function value at $0$ is:
$$f(0) = \frac{1}{2}\|M-U\Sigma V^T\|_F^2.$$
However, $f(0)$ need not be computed as we are only interested in
$f'(0)$ and $f''(0)$ for obtaining $\tilde{t}^*$.

Second, since $f'(t) = \tr \Big[ h(t)^T\cdot h'(t) \Big]$ we have
$$f'(0) = \tr \Big[ h(0)^T\cdot h'(0) \Big],$$
where
\begin{eqnarray*}
  h(0) &=&  M-U(0)\Sigma(0)V(0)^T,\\
  h'(0) &=& -U'(0)\Sigma(0)V(0)^T - U(0)\Sigma'(0)V(0)^T \\
  &&- U(0)\Sigma(0)V'(0)^T.
\end{eqnarray*}
Given $ U(0)=U, \Sigma(0)=\Sigma, V(0)=V, U'(0)=-P_UU,
\\ \Sigma'(0)=-P_\Sigma, V'(0)=-P_VV$, we have
$$h'(0) = P_U(U\Sigma V^T) + U P_\Sigma V^T - (U\Sigma V^T)P_V.$$

Finally, we come to the second-order derivative:
$$f''(0) = \tr\Big[ h'(0)^T\cdot h'(0) + h(0)^T\cdot h''(0)\Big],$$
where
\begin{equation*}
\begin{array}{rl}
    &h''(0) \\
    = &-U''(0)\Sigma(0)V(0)^T - U'(0)\Sigma'(0)V(0)^T - U'(0)\Sigma(0)V'(0)^T\\
    &- U'(0)\Sigma'(0)V(0)^T - U(0)\Sigma''(0)V(0)^T - U(0)\Sigma'(0)V'(0)^T\\
    &- U'(0)\Sigma(0)V'(0)^T - U(0)\Sigma'(0)V'(0)^T - U(0)\Sigma(0)V''(0)^T,
\end{array}
\end{equation*}
with $U''(0)=P_U^2U$, $\Sigma''(0)=0$ and $V''(0)=P_V^2V$.

Thus we have
\begin{equation*}\label{Eq:h''0}
\begin{array}{rl}
   h''(0) =& -P_U^2U\Sigma V^T - 2\cdot P_U U P_\Sigma V^T + 2\cdot P_UU\Sigma V^TP_V \\
  &+ 2\cdot U P_\Sigma V^TP_V - U\Sigma V^TP_V^2\\
  =&-P_U Z + ZP_V,
\end{array}
\end{equation*}
where $Z=h'(0)+U P_\Sigma V^T$.

\bibliographystyle{spbasic}

\end{document}